\documentclass[10pt,twocolumn,letterpaper]{article}

\usepackage[pagenumbers]{cvpr} 

\newcommand\blfootnote[1]{%
  \begingroup
  \renewcommand\thefootnote{}\footnote{#1}%
  \addtocounter{footnote}{-1}%
  \endgroup
}
\def\fk{\mathbf{F}_k}
\def\ft{\mathbf{F}_t}
\def\fs{\mathbf{F}_s}
\def\It{\mathbf{I}_t}
\def\Is{\mathbf{I}_s}

\def\vV{\mathbf{V}}
\def\wq{\mathbf{W}_Q}
\def\wk{\mathbf{W}_K}
\def\wv{\mathbf{W}_V}
\def\vA{\mathbf{A}}

\def\Ks{\mathbf{K}_s}
\def\Kt{\mathbf{K}_t}
\def\Ct{\mathbf{C}_t}

\def\fkp{\mathbf{F}_k'}

\def\ftp{\mathbf{F}_t'}

\def\xt{\mathbf{X}_t}
\def\ks{\mathbf{K}_s}
\def\kt{\mathbf{K}_t}

\def\pk{\mathbf{P}_k}

\def\vQ{\mathbf{Q}}

\definecolor{cvprblue}{rgb}{0.21,0.49,0.74}
\usepackage[pagebackref,breaklinks,colorlinks,allcolors=cvprblue]{hyperref}

\def\paperID{} 
\def\confName{CVPR}
\def\confYear{2026}

\title{UniCorrn: Unified Correspondence Transformer Across 2D and 3D}

\author{%
  Prajnan Goswami$^{1*}$\quad
  Tianye Ding$^{1*}$\quad
  Feng Liu$^2$ \quad
  Huaizu Jiang$^1$
  \\
  $^1$Northeastern University \quad
  $^2$Adobe Research \quad
  \\
}

\usepackage[normalem]{ulem}
\usepackage{multirow}
\usepackage{pifont}
\usepackage{colortbl}
\usepackage{xcolor}
\usepackage{overpic}

\begin{document}

\newcommand{\cmark}{\ding{51}}
\newcommand{\xmark}{\ding{55}}
\newcommand{\RNum}[1]{\MakeUppercase{\romannumeral #1\relax}}

\twocolumn[{
    \renewcommand\twocolumn[1][]{#1}%
    \maketitle
    \centering
    \vspace{-10mm}
    \includegraphics[width=0.96\linewidth]{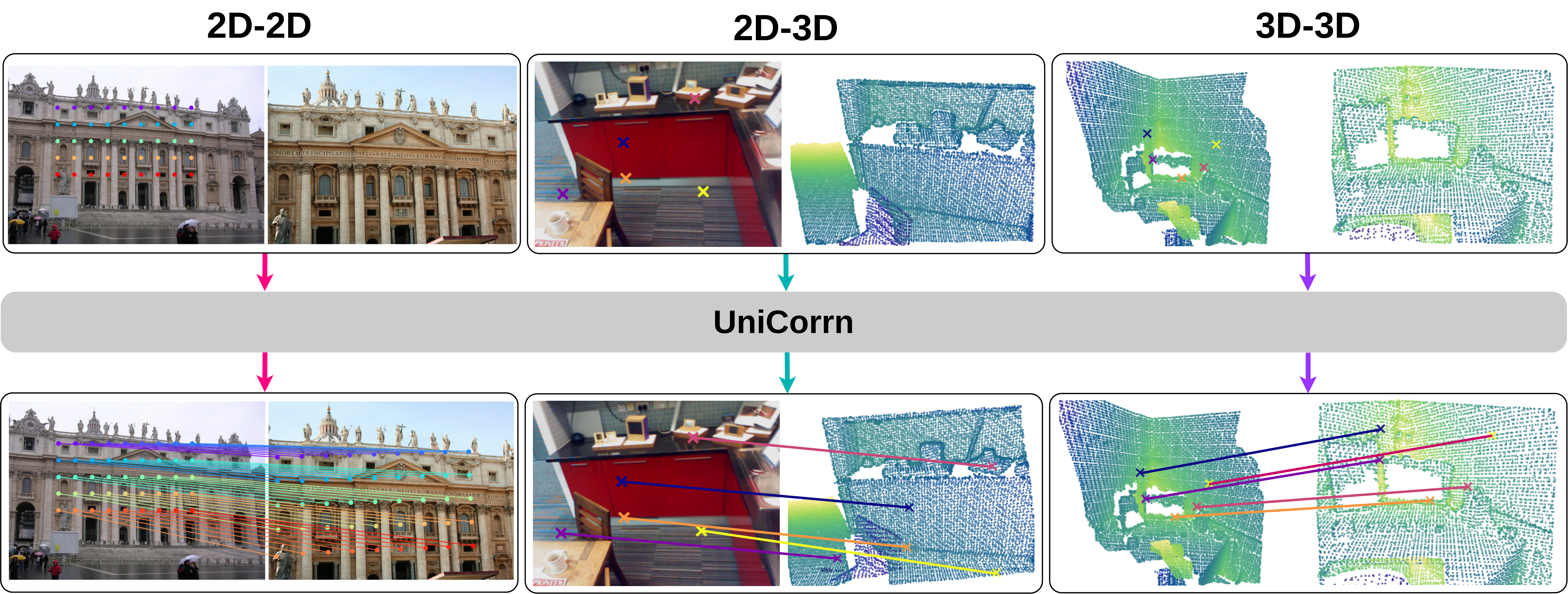}
        \captionof{figure}{
            \textbf{UniCorrn} is a unified correspondence transformer that can find correspondences of keypoints of interest across 2D and 3D.
        }
        \vspace{10pt}
    \label{fig:teaser}
}]

\blfootnote{$^*$ Equal contribution}
\begin{abstract}

Visual correspondence across image-to-image (2D-2D), image-to-point cloud 
(2D-3D), and point cloud-to-point cloud (3D-3D) geometric matching forms 
the foundation for numerous 3D vision tasks. Despite sharing a similar problem structure, 
current methods use task-specific designs with separate models for each 
modality combination. We present UniCorrn, the first correspondence model 
with shared weights that unifies geometric matching across all three tasks. 
Our key insight is that Transformer attention naturally captures cross-modal feature similarity. 
We propose a dual-stream decoder that maintains 
separate appearance and positional feature streams. This design enables 
end-to-end learning through stack-able layers while supporting flexible 
query-based correspondence estimation across heterogeneous modalities. Our 
architecture employs modality-specific backbones followed by shared encoder 
and decoder components, trained jointly on diverse data combining pseudo point 
clouds from depth maps with real 3D correspondence annotations. UniCorrn 
achieves competitive performance on 2D-2D matching and surpasses prior 
state-of-the-art by 8\% on 7Scenes (2D-3D) and 10\% on 3DLoMatch (3D-3D) in registration recall. 
Project website: \href{https://neu-vi.github.io/UniCorrn/}{\textcolor{magenta}{neu-vi.github.io/UniCorrn/}}

\end{abstract}

\section{Introduction}

Visual correspondence, the task of finding matching features across different observations of the same  scene, plays a fundamental role in 3D computer vision.
Geometric keypoint matching can be categorized into three types: image-to-image (2D-2D), image-to-point cloud (2D-3D), and point cloud-to-point cloud (3D-3D) matching, as shown in Figure~\ref{fig:teaser}. 
These inter-modal and intra-modal keypoint matches form the foundation for various downstream applications, including point cloud registration~\cite{direct_superpoints_matching}, camera pose estimation~\cite{Reloc3r}, structure from motion and SLAM~\cite{vggsfm,OV2SLAM}.

Although significant advances have been made in solving various forms of 
visual correspondence problems, different specialist models maintain different 
task-specific designs~\cite{DKM,LoFTR,GeoTransformer,Predator,2D3D-MATR,Bridge-2D3D} despite the similar nature of the problem across 2D and 3D domains. 
While some works have explored unified matching within the 2D image 
domain~\cite{choy2016universal,truong2020glu,zheng2023rgm,%
he2025MatchAnything,xue2025matcha,UniFlowMatch_2025}, no solution exists for 
geometric correspondence across 2D and 3D modalities. 
In this paper, we ask: 
\emph{is it possible to approach geometric matching across 2D and 3D 
modalities using a unified model?} A unified correspondence model not only 
represents a grand scientific pursuit toward general-purpose visual 
perception, but also has the potential to enable seamless cross-modal 
reconstruction pipelines, reduce engineering complexity, and facilitate 
learning of shared geometric priors across modalities through joint training.

Addressing this question requires overcoming fundamental methodology 
limitations in existing 2D unification approaches that prevent their extension 
to 3D domains. These efforts can be broadly grouped into three categories, 
each with distinct architectural constraints. First, cost volume-based 
methods~\cite{truong2020glu,he2025MatchAnything,zheng2023rgm} capture feature 
similarity within local ranges to ensure efficiency, from which coarse-to-fine 
estimations are performed via image pyramids or recurrent networks. However, 
the fixed depth of pyramids or sequential nature of recurrent operations 
limits their representational capacity, making them unsuitable for handling 
the sparse and irregular structure of 3D point clouds where correspondences 
may span large spatial distances. Second, nearest-neighbor (NN) search 
methods~\cite{choy2016universal,xue2025matcha} match dense feature 
descriptors, but NN search can only be performed once and cannot be 
incorporated into stacked neural network layers for end-to-end training. 
This prevents iterative feature refinement necessary for learning robust 
cross-modality alignments between heterogeneous 2D and 3D representations. 
Third, while direct regression approaches~\cite{UniFlowMatch_2025} fuse image 
features with transformers and directly regress dense pixel displacements, 
our experiments show that direct regression struggles in 2D-3D and 3D-3D 
settings where explicit geometric reasoning about 3D structure is essential 
for accurate correspondence estimation. These limitations motivate our 
approach: we need a matching mechanism that (1) supports end-to-end learning 
through stackable layers, (2) handles irregular structures across modalities, 
and (3) enables iterative geometric refinement of correspondence estimation.

In this paper, we present \textbf{UniCorrn}, a unified correspondence model based on 
the Transformer architecture~\cite{vaswani2017attention} that addresses 
geometric matching tasks across 2D-2D, 2D-3D, and 3D-3D modalities. Our key 
insight is that attention mechanism in Transformers naturally captures feature 
similarity, which is the essence of correspondence across all modalities. 
To effectively leverage this property for cross-modal matching, we develop 
a novel dual-stream attention mechanism in our matching decoder, where we 
maintain separate residual streams for appearance and positional features. 
These streams are combined to compute attention maps, based on which both 
appearance and positional features are updated independently. This design 
enables us to regress matching keypoint locations from attention-modulated 
positional encodings while supporting end-to-end learning through stacked 
Transformer layers. Crucially, our model employs modality-specific backbones 
followed by a shared feature fusion encoder and matching decoder with 
identical weights across all input modality combinations. This weight-sharing 
design, as opposed to training separate models for each task, enables joint 
learning of geometric priors across 2D and 3D domains. Given source keypoints 
of interest, our model directly decodes their corresponding locations in the 
target modality, providing a flexible query-based interface for correspondence 
estimation.

Inspired by recent foundation models for computer vision~\cite{dino-v2,kirillov2023segment,%
CroCov2,MASt3R,VGGT}, we train our unified model on diverse correspondence 
data across modalities. 
We leverage pretrained CroCo v2~\cite{CroCov2} for 
image feature extraction and build our correspondence model with $600$M 
parameters, enabling rich representation learning while maintaining 
computational efficiency. 
A key challenge is the scarcity of training data for 
2D-3D and 3D-3D correspondences compared to abundant 2D-2D image pairs. To 
address this, we combine pseudo point-cloud data derived from depth maps used 
in DUSt3R training~\cite{DUSt3R} with smaller amounts of high-quality 3D 
correspondence annotations~\cite{LCD,3dmatch,bai2020d3feat}. This mixed 
training strategy enables our model to learn robust geometric priors across 
modalities. Experimental results show that UniCorrn achieves competitive 
performance in 2D-2D matching and surpasses existing methods in 2D-3D by 
8\% and 3D-3D by 10\% in registration recall on standard  
benchmarks. Extensive ablation studies further validate the effectiveness of 
our model design, especially the dual-stream matching decoder.

In summary, we make three major contributions:
\begin{itemize}[itemsep=0pt,topsep=0pt,leftmargin=18pt]%
    \item We present UniCorrn, the first unified correspondence model with shared weights for 
    geometric matching across 2D-2D, 2D-3D, and 3D-3D modalities.
    
    \item We propose a novel dual-stream Transformer decoder that decouples 
    appearance and positional features, enabling stackable layers for 
    correspondence matching.

    \item We achieve state-of-the-art results on 7Scenes~\cite{7Scenes} (2D-3D) and 3DLoMatch~\cite{3dmatch} (3D-3D) correspondence, 
    surpassing prior methods by 8\% and 10\% in registration recall respectively, 
    while maintaining competitive performance on 2D-2D matching.

\end{itemize}

\section{Related Work}

\begin{figure*}[!th]
  \centering
   \includegraphics[width=\linewidth, trim=50 100 70 40, clip]{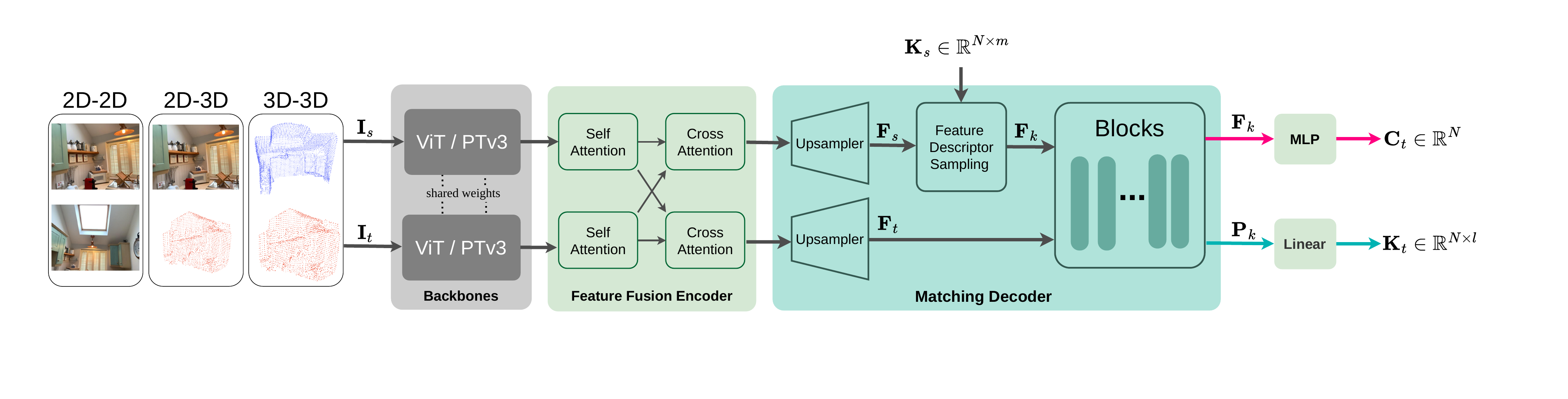}
  \caption{\textbf{Illustration of the overall architecture design.} Our model consists of 
   four main modules: (1) modality-specific backbone, (2) feature fusion encoder, (3) matching decoder,
  and (4) modality-specific prediction heads. Details of each module can be found in
  Sec.~\ref{subsec: network architecture}.
  }

\label{fig:full_arch}
\vspace{-3mm}
\end{figure*}

\textbf{Image to image (2D-2D).}
In 2D-2D matching, learning-based methods include keypoint detection, feature description extraction, and feature matching~\cite{lowe_scaleinv,SuperPoint,SuperGlue,R2D2,DISK}. 
Recent methods~\cite{LoFTR,ASpanFormer,DKM,RoMa,MASt3R} have replaced keypoint detection with detection-free approach. These methods perform dense matching by feature warping for spatial alignment~\cite{DKM,PATS,truong2020glu,RoMa, he2025MatchAnything} or by computing similarity between features followed by nearest neighbor search~\cite{LoFTR,Efficient_LoFTR,ASTR, xue2025matcha,MASt3R}. COTR~\cite{COTR} and VGGT~\cite{VGGT} allow users to query keypoints on one image and directly estimate the correspondences on the other. The query method is commonly used in video key-point tracking tasks~\cite{TAP-Vid,TAPIR,CoTracker3,VGGT}, however, it is underexplored for geometric matching.

\noindent\textbf{Image to point cloud (2D-3D).}
To predict correspondences between images and point clouds, seminal deep learning methods~\cite{2D3D-Matchnet_outdoor, LCD} directly match pairs of learned local image patch and local point cloud volume descriptors with a distance metric. For dense per-pixel/per-point correspondences, DeepI2P~\cite{DeepI2P_outdoor} classifies whether each point in the point cloud lies within or beyond the camera frustum. Recent work~\cite{P2-net,CoFiI2P_outdoor,2D3D-MATR,Bridge-2D3D} achieves better matching with circle loss~\cite{CirleLoss} and adopts a coarse-to-fine matching strategy. FreeReg~\cite{FreeReg} and DiffReg~\cite{Diff-Reg} incorporate diffusion~\cite{DDPM} into the matching pipeline, improving cross-modality matching at the cost of diffusion sampling time. 

\noindent\textbf{Point cloud to point cloud (3D-3D).}
Learning-based 3D-3D methods can be broadly classified into matching with 3D local descriptors and point-cloud registration. Early work~\cite{3dmatch, bai2020d3feat, Predator} extracts
local 3D patch descriptors and estimates 3D-3D correspondences by computing per-point overlap and matching scores. Recent state-of-the-art methods~\cite{Regtr, GeoTransformer, Li_2022_CVPR, roltr, peal3d} use Transformer~\cite{vaswani2017attention} with cross-attention to enhance 3D superpoint features. These methods directly supervise the 
model on rigid SE(3) transformation to avoid the high computation cost of matching dense feature descriptors.

\noindent\textbf{Unified correspondence models.}
Unified correspondence models aims to solve more than one correspondence tasks. UCN~\cite{choy2016universal} supervises CNN feature maps between image pairs with contrastive loss and
uses nearest neighbor search to estimate geometric and semantic correspondences. Glu-Net~\cite{truong2020glu} unifies geometric, semantic, and optical flow by computing similarity for across feature pyramids. RGM~\cite{zheng2023rgm} proposes a two-stage model with iterative refinement for dense flow and sparse geometric matching. MatchAnything~\cite{he2025MatchAnything}
supervises existing 2D-2D matching methods~\cite{RoMa, Efficient_LoFTR} with large-scale data containing multiple imaging modalities such as thermal, tomography, histology, etc. MATCHA~\cite{xue2025matcha} incorporates features extracted from foundational models, Stable Diffusion~\cite{sd_diffusion} and DINOv2~\cite{dino-v2}, to unify matching across geometric, semantic and temporal keypoint tracking. UFM~\cite{UniFlowMatch_2025} fuses image features with a global attention Transformer and directly regresses the enhanced features for 2D-2D geometric and temporal matching.

\section{Method}
The input to our model consists of $\Is, \It$ and a list of keypoints of interest $\Ks \in \mathbb{R}^{N \times m}$ in $\Is$, where $\Is$ and $\It$ represent the input source and target modalities, respectively.
$m \in \{2,3\}$ indicates the modality dimension depending on the task specification.
The source and target pair $\Is, \It$ can be formed between image-to-image (2D-2D), image-to-point (2D-3D), and point-to-point (3D-3D). 
The keypoints $\Ks$ can be either from a detector~\cite{SIFT, SuperPoint, Aliked} or sampled from an equally spaced grid.
The output  are a set of matching keypoints $\Kt \in \mathbb{R}^{N \times l}$ in the target $\It$ with $l 
\in \{2,3\}$ depending on the modality of $\It$, and confidence scores $\Ct \in \mathbb{R}^{N}$.
The confidence scores quantifies the model's uncertainty in matching keypoints in challenging areas like occluded regions, translucent objects, sky, etc.

\subsection{Network Architecture}
\label{subsec: network architecture}
We design a unified correspondence model based on Transformer~\cite{vaswani2017attention} following recent large-scale models~\cite{CroCov2, DUSt3R, MASt3R, VGGT} in 3D computer vision. As shown in Fig.~\ref{fig:full_arch}, our model consists of four main modules: (1) modality-specific backbone, (2) feature fusion encoder, (3) matching decoder, and (4) modality-specific prediction head.

\noindent\textbf{Modality-specific backbones.}
We use separate feature extractors for images and point clouds. 
Specifically, we use a ViT~\cite{ViT} for 2D images and Point Transformer v3 (PTv3)~\cite{PTv3} for 3D point clouds. ViT and PTv3 have shown state-of-the-art performance in various computer vision tasks,
thereby making them a good choice for our unified correspondence Transformer model. The backbone weights are shared in a Siamese manner when both source $\Is$ and target $\It$ belong to the same modality. Rotary position embeddings~\cite{RoFormer_RoPE} are used to encode relative positional information for both image and point cloud tokens.

\noindent\textbf{Feature fusion encoder.} 
We do not assume any modality specifics at this stage. The feature fusion encoder takes as input the modality-specific features of $\Is$ and $\It$.
We use a generic design here following existing matching frameworks~\cite{LoFTR, direct_superpoints_matching, 2D3D-MATR, CroCov2} to allow information exchange between the input via cross-attention. Each encoder block has alternating layers of self-attention, where each token attends to all tokens of the same input, and cross-attention, where each token attends to all tokens of the other input.

\noindent\textbf{Matching decoder.}
Our main contribution is the Transformer-based matching decoder, where we propose a novel dual-stream attention module for keypoints matching.
First, the fused image features from the output of the feature fusion encoder are upsampled using an MLP with Pixel Shuffle~\cite{PixelShuffle}, and a PTv3’s learned upsampler~\cite{PTv3} is used for fused point features.
The upsampled features corresponding to the \emph{source} and \emph{target} inputs are indicated by $\fs$ and $\ft$, respectively.
Along with keypoints-of-interest of the \emph{source} input $\Ks$, they are processed by a set of dual-stream Transformer layers, outputting a positional embedding $\pk$ and appearance feature $\fk$, which contain useful feature representations for regressing correspondences in the target and estimating the uncertainties, respectively.
We will introduce this module with more details in Sec.~\ref{subsec: matching decoder}.

\noindent\textbf{Prediction heads.}
The positional embedding $\pk$ from the matching decoder is fed into modality-specific linear layers to regress the  2D or 3D coordinates $\Kt$ of corresponding keypoints. And a shared MLP takes as input the updated keypoint features $\fk$ from the matching decoder and predicts confidence scores $\Ct$ for the correspondences.

\subsection{Matching Decoder}
\label{subsec: matching decoder}

\noindent\textbf{Encoding keypoints of interest.}
Given the keypoints of interest $\ks$, we obtain keypoint descriptors, denoted as $\fk$, from $\fs$ using bilinear interpolation if $\Is$ is an image. 
If it is a point cloud, a shared Gaussian distribution with a learnable $\sigma$ is applied to produce a single weighted feature vector from k-nearest features of each 3D keypoint. 

\noindent\textbf{Capturing similarity via attention in Transformer.}
The essence of various correspondence tasks by definition is to capture the similarity between $\fk$ and $\ft$.
Our core insight of designing the matching decoder is that the attention matrix in a Transformer layer captures the \textbf{matching cost} between the input pair, \ie, correlation between two inputs in existing correspondence tasks~\cite{FlowNet, LoFTR, 2D3D-MATR, direct_superpoints_matching}.
Specifically, we first compute position-augmented features
\begin{align}
    \fkp = \texttt{RoPE}(\fk\wq, \kt),~
    \ftp = \texttt{RoPE}(\ft\wk,\xt),\notag
\end{align}

\noindent where \texttt{RoPE} represents rotary position embedding~\cite{RoFormer_RoPE}. $\xt$ are the coordinates of the tokens in $\ft$.
$\kt$ are the estimated coordinates of corresponding keypoints according to Eq.(\ref{eq:bijective mapping}) below.
$\wq$ and $\wk$ are the weight matrices associated with the query and key in Transformer, respectively. 
The attention matrix is then computed as
\begin{align}
    \vA = \texttt{Softmax}\left(\frac{\fkp\ftp^T}{\sqrt{D}}\right),
    \label{eq:attention_toy}
\end{align}
where $D$ is the feature dimension.
This attention matrix $\vA$ is similar to the normalized version of the learnable cost volume studied in~\cite{xiao20learnable}.
In an ideal case with perfect similarity scores, each row in $\vA$ 
is a one-hot vector, where the position of 1 corresponds to the correct matching keypoint.
A Transformer layer then works  as\footnote{We omit module such as FFN, LayerNorm, here for brevity.}
\begin{equation}
    \vQ = \vA \vV + \vQ.
    \label{eq:plain_attention}
\end{equation}
Here $\vQ$ and $\vV$ denote the query and value vector in a Transformer in general. 
If we set $\vV$ to the \emph{absolute positional encoding} of every pixel in $\It$, the updated query $\vQ$ contains the positional encoding of the correct corresponding pixels for every keypoint in $\Ks$, from which we can regress the coordinates.
\emph{The readers are highly encouraged to check an illustration provided in the supplementary material.}

\begin{figure}[t]
  \centering
   \includegraphics[width=\linewidth, trim=0 5 0 10, clip]{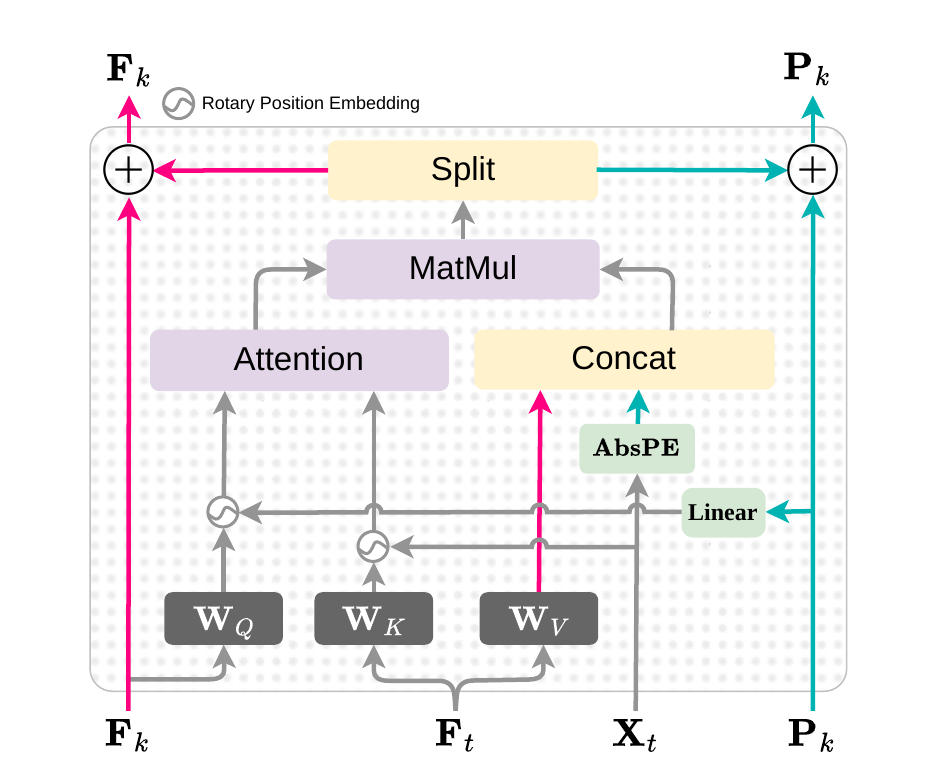}

  \caption{\textbf{Dual-stream attention with a single attention matrix (matching cost).} The appearance and position features are concatenated along the channel dimension to process them in parallel. After applying attention, the output is split to update the corresponding \emph{appearance} $\fk$ and \emph{positional} $\pk$ residual streams.}
\label{fig:two-stream_ca}
\vspace{-5mm}
\end{figure}

\noindent\textbf{Dual-stream Transformer design.}
The power of the Transformer design lies in that multiple Transformer layers can be stacked to refine the input, where the output of one layer will be used as input to the next layer.
It is not feasible, however, to directly stack multiple Transformer layers introduced in Eq.(\ref{eq:plain_attention})
as the output $\vQ$ consists of \emph{positional encoding only}, which cannot be used to match the appearance features $\fk$ in subsequent Transformer layers.  To overcome this issue, we propose a dual-stream design for a Transformer layer by separating the apperance and positional embeddings.
An illustration is shown in Fig.~\ref{fig:two-stream_ca}.
Our ablation study shows that it works better than other instantiations of Transformer for visual correspondences~\cite{COTR,UniFlowMatch_2025}.
First, $\fk$ is updated in the first stream as
\begin{align}
    \fk = \vA (\wv\ft) + \fk.
    \label{eq:appearance_stream}
\end{align}

Second, we introduce the other stream with a positional embedding $\pk$, defined as
\begin{align}
    \pk=\vA (\texttt{AbsPE}(\xt)) + \pk,
    \label{eq: position stream}
\end{align}
where $\texttt{AbsPE}(\xt) = \mathbf{W_p}\xt + \mathbf{b_p}$ indicates a learned \emph{bijective} absolute positional encoding with parameters $\mathbf{W_p}$ and $\mathbf{b_p}$. 
The positional embedding $\pk\in\mathbb{R}^{N\times D}$ is initialized with zeros.
Similar to how the appearance features $\fk$ is updated in Eq.(\ref{eq:appearance_stream}), here the positional embeddings is updated separately.
Note that both the appearance features and positional embeddings will still be combined to compute the attention matrix $\vA$.
In practice, we replace the vanilla attention in Eq.(\ref{eq:attention_toy})  with a Gaussian variant 
\begin{align}
    \vA = \texttt{Softmax}\left(-\frac{\texttt{Pair\_L2}(\fkp,  \ftp)}{D}\right),
    \label{eq:attention_gaussian}
\end{align}
where \texttt{Pair\_L2} computes the pairwise L2 distance.
The vanilla attention matrix can be interpreted as using a linear kernel to compute the feature similarities between query and key, which only capture linear correlations and are sensitive to the scales of the magnitude of features.
Similar to matching through descriptors~\cite{lowe2004distinctive, bo2010kernel}, we use a Gaussian kernel to capture the non-linear complex correlations. 
Experimental results show that it works better.

\noindent\textbf{Regressing the coordinates of correspondences.}
With the output positional embedding $\pk$, we can directly regress the coordinates of the corresponding keypoints $\kt$ using a linear layer as 

\begin{equation}
    \mathbf{\kt} = \mathbf{W_p^+}(\pk - \mathbf{b_p}),
    \label{eq:bijective mapping}
\end{equation} 
where $\mathbf{W_p^+}$ is the Moore–Penrose inverse~\cite{moore1920reciprocal, penrose1955generalized} of $\mathbf{W_p}$. 
We also estimate pixel-wise confidence scores $\Ct$ for the output correspondences using a shared MLP for all modalities that takes $\fk$ as input.

\noindent\textbf{Stacking dual-stream Transformer layers.} 
By decomposing the appearance features $\fk$ and positional embeddings $\pk$ into two separate streams, we can stack multiple such Transformer layers together, where the updated $\fk$ and $\pk$ of the current layer will be fed into the subsequent layer.
As a result, by stacking multiple layers together, both $\fk$ and $\pk$ will be gradually refined, leading to more accurate attention matrices and thus more accurate correspondence estimation.

\subsection{Training Objective}
\label{subsec:training objective}

Our model is supervised with a loss for jointly training all three tasks 
\begin{equation}
    \mathcal{L}_{total} = \mathcal{L}_{2d2d} + \mathcal{L}_{2d3d} + \mathcal{L}_{3d3d}.
\end{equation}
For each task, we consider the following three objectives
\begin{equation}
    \mathcal{L}_{task} = \mathcal{L}_{conf} + \mathcal{L}_{aux} + \beta\mathcal{L}_{desc},
\end{equation}
where $task\in\{2d2d, 2d3d, 3d3d\}$. $\beta$ is a weight to balance the loss terms. 
The three losses are introduced below.

\noindent\textbf{Confidence-aware L1 Loss.} 
The model is directly supervised with the error of the predicted keypoints using L1 loss. 
To quantify the uncertainty of predictions in parts of the input such as sky, occluded objects, etc.,
we incorporate the confidence-aware loss adapted from MASt3R~\cite{MASt3R}. 
\begin{equation}
    \mathcal{L}_{conf} = \frac{1}{N}\sum_{i=1}^N \Ct(i)\big\|{\Kt}(i) - {\mathbf{\bar{K}_t}}(i)\big\|_1 - \alpha \text{log}\Ct(i),\notag
\end{equation}
where $\mathbf{\bar{K}_t}$ denote the ground-truth coordinates of corresponding keypoints.
$\alpha$ is a regularization strength. 

\noindent\textbf{Contrastive loss.} 
We supervise the input features to have one-to-one correspondences using the InfoNCE loss~\cite{infoNCE}, which helps improve the attention matrix. Specifically, we use ground-truth correspondence pairs to extract feature descriptors $\fs^{desc}$ and $\ft^{desc}$ from the upsampled fused features $\fs$ and $\ft$, respectively. We then compute the InfoNCE loss ($\mathcal{L}_c$) between these descriptors, and also apply the same loss to the output of the matching decoder $\fk$ and the extracted \emph{target} features $\ft^{desc}$.
\begin{equation}
    \mathcal{L}_{desc} = \mathcal{L}_{c}(\fs^{desc}, \ft^{desc}) 
                       + \mathcal{L}_{c}(\fk, \ft^{desc}).
     \label{eq:total_infonce}
\end{equation}
Detailes are provided in the supplementary material.
\def\ktl{\mathbf{K_t^{(\mathnormal{l})}}}

\noindent\textbf{Auxiliary supervision.} 
We also estimate the coordinates of corresponding keypoints at each of the matching decoder layer to provide auxiliary supervision.
Let $\ktl$ denote the estimated coordinates at the $l$-th decoder layer.
The auxiliary loss is defined as
\begin{equation}
    \mathcal{L}_{aux} = \sum_{l=1}^L \gamma^{L - l} \frac{1}{N}\sum_{i=1}^N \big\|\ktl(i) - \mathbf{\bar{K}_t}(i)\big\|_1,
\end{equation}
where $L$ is the total number of matching decoder layers and $\gamma$ is a coefficient set to 0.9.

\section{Experiments}

\subsection{Setup}
\label{subsec: setup}

\noindent\textbf{Datasets.}
We train our unified model on the 2D-2D task with 7 datasets: ARKitScenes~\cite{arkitscenes}, 
BlendedMVS~\cite{BlendedMVS}, CO3D-v2~\cite{CO3D}, MegaDepth~\cite{megadepth}, StaticThings3D~\cite{things3d},
ScanNet++~\cite{scannetpp} and Waymo~\cite{waymo}. 
For the 2D-3D task, we use 7Scenes~\cite{7Scenes} and RGB-D Scenes v2~\cite{RGB-DScenesV2}.
And finally for the 3D-3D, 3DMatch~\cite{3dmatch} and ModelNet~\cite{modelnet} are used. 
We complement the 2D-3D and 3D-3D datasets with pseudo data generated from the dense depth maps of ScanNet++~\cite{scannetpp} and ARKiTScenes~\cite{arkitscenes}.

\begin{table}
\small
\setlength{\tabcolsep}{2pt}
\centering
\renewcommand{\midrule}{\noalign{\vskip 3pt\hrule height 0.8pt\vskip 3pt}}
\caption{\textbf{Ablation of different matching paradigms on single task small-scale experiments.} The top two methods represent \textit{dense matching} design and the bottom four rows represent \textit{keypoint queryable} design.}
\vspace{-2mm}
\label{table:ablation_matching_methods}
\begin{tabular}{l|cc|cc|cc} 
\toprule
\multirow{2}{*}{\begin{tabular}[c]{@{}l@{}}Matching \\Paradigm \end{tabular}}
                       & \multicolumn{2}{c|}{\begin{tabular}[c]{@{}c@{}}MegaDepth \\(2D-2D) \end{tabular}} 
                       & \multicolumn{2}{c|}{\begin{tabular}[c]{@{}c@{}}7Scenes \\(2D-3D) \end{tabular}} 
                       & \multicolumn{2}{c}{\begin{tabular}[c]{@{}c@{}}3DMatch \\ (3D-3D) \end{tabular}} \\ 
                       
\cmidrule{2-3}\cmidrule{4-5}\cmidrule{6-7}
                       & $5^\circ \uparrow$  
                       & $10^\circ \uparrow$
                       & IR  $\uparrow$
                       & RR  $\uparrow$ 
                       & IR  $\uparrow$
                       & RR  $\uparrow$ \\ 
\midrule		

Nearest neighbor                                & 49.8      & \textbf{67.1}            & 42.0             &    63.4         & 24.8              & 87.5 \\
Global matching                                 & 48.8      & 64.9                     & 65.2             &    75.8         & \textbf{93.1}     & 96.5 \\

\cmidrule{1-7}
    
Regression                                      & 0.2       & 1.5	                    & 10.0            &   17.0           & 5.7    & 18.2             \\                       
Sequence concatenation                          & 10.3      & 22.1	                    & 12.8            &   21.1           & 2.4    & 7.3              \\
Ours     &  \textbf{50.6}      & \textbf{67.1}	& \textbf{66.3}   &   \textbf{77.8}  & 92.8   & \textbf{96.9}    \\

\bottomrule
\end{tabular}
\vspace{-2mm}
\end{table}

\noindent\textbf{Evaluation protocols.}
To measure the performance of 2D-2D, we report the \emph{Area Under Curve} (AUC) of the 
relative pose errors at $5^\circ$, $10^\circ$ and $20^\circ$ degree thresholds following
the evaluation protocol in ~\cite{SuperGlue, LoFTR, ASpanFormer}.
The pose error is defined as the maximum of angular errors in rotation and translation. 
For 2D-3D and 3D-3D, we follow the evaluation protocol in ~\cite{2D3D-MATR} and ~\cite{Predator, Regtr}, respectively.
Specifically, we report: 
(1) \emph{Inlier Ratio} (IR), the ratio of pixel-to-point 
or point-to-point matches whose 3D distance is below a 
certain threshold over all putative matches;
(2) \emph{Feature Matching Recall} (FMR), the ratio of 
2D-2D or 3D-3D pairs whose IR is above a certain threshold;
(3) \emph{Registration Recall} (RR), the ratio of 2D-3D or 
3D-3D pairs whose RMSE is below a certain threshold.
Additionally, we report the registration pose errors as \emph{Relative Rotation Error} (RRE) 
and \emph{Relative Translation Error} (RTE) on 3DMatch~\cite{3dmatch} and 
ModelNet~\cite{modelnet}. 

\noindent\textbf{Implementation details.}
We train two models with different capacities. A small-scale model is employed for the ablation study in Section~\ref{subsec: ablation study}, and a scaled up version for the benchmark in Section~\ref{subsec: benchmarks}. The large-scale model is trained in two stages. Complete architectural details and training schemes are available in the supplementary materials.

\begin{table}
\small
\setlength{\tabcolsep}{4pt}
\centering
\renewcommand{\midrule}{\noalign{\vskip 3pt\hrule height 0.8pt\vskip 3pt}}

\caption{\textbf{Ablation of different design choices}. We analyze the impact of our contributions in the query matching decoder with detailed explanations provided in Section~\ref{subsec: ablation study}. $D$ and $H$ refers to embedding dimensions and number of attention heads. }
\label{table:ablation_two_stream}
\begin{tabular}{l|ccc} 
\toprule
\multirow{2}{*}{\begin{tabular}[c]{@{}l@{}}Setup \end{tabular}}
                       & \multicolumn{3}{c}{MegaDepth-1500}  \\
                       
\cmidrule{2-4}
                       & $5^\circ \uparrow$  
                       & $10^\circ \uparrow$  
                       & $20^\circ \uparrow$ \\

\midrule 
\label{case:I}\textcolor{black}{\RNum{1} (Baseline):}  $D=256, H=16$    & 36.4	             &  53.9	          &  68.9    \\  
\label{case:II}\RNum{2}: \RNum {1}, Gaussian Attention                         & 37.3	             &  54.7              &  69.5    \\    
\label{case:III}\RNum{3}: \RNum {2}, 800 Keypoint Queries                      & 38.2	             &  55.8              &  70.7    \\ 
\label{case:IV}\RNum{4}: \RNum {3}, $D=256, H=1$                         & 39.5               &  57.0              &  71.3    \\ 
\label{case:V}\RNum{5}: \RNum {4}, Contrastive Loss                            & 43.9      & 60.7     & 74.5     \\    
\label{case:VI}\RNum{6}: \RNum {5}, $4\times$ Upscale, $D=64, H=1$       & 48.5	             &  65.1              &  77.9    \\     
	 	 
\textcolor{black}{\label{case:VII}\textbf{\RNum{7} (Final)}:} \RNum {6}, $D=256, H=1$   & \textbf{50.6}       &  \textbf{67.1}     &  \textbf{79.6}    \\ 	  
\bottomrule
\end{tabular}
\vspace{-6mm}
\end{table}

\subsection{Ablation Study}
\label{subsec: ablation study}

\begin{figure}
    \centering
    \includegraphics[width=0.7\linewidth]{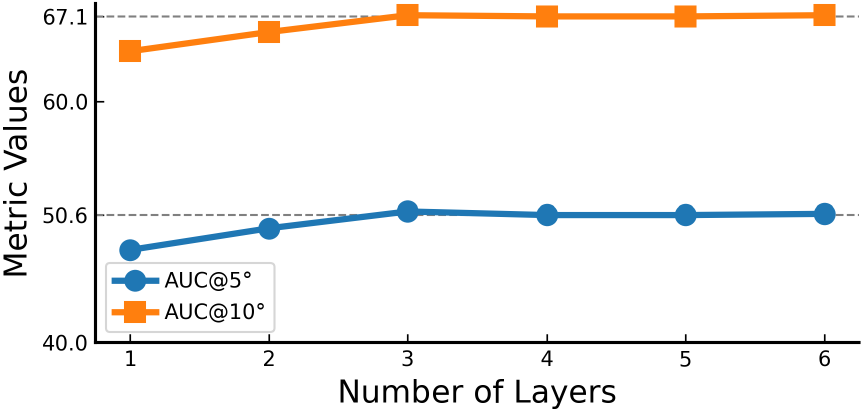}
    \includegraphics[width=0.7\linewidth]{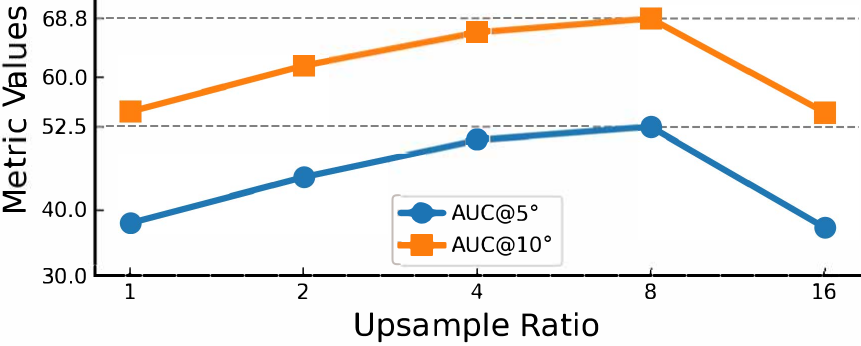}
    \vspace{-2pt}
    \caption{\textbf{Top:} AUC \vs number of matching decoder layers. \textbf{Bottom}: AUC \vs feature upsampling ratio. 
    The results are obtained on the MegaDepth-1500 dataset.
    }
    \label{fig:ablation plots}
    \vspace{-4mm}
\end{figure}

\noindent\textbf{Methods for estimating correspondences.}
We first study the effectiveness of our proposed dual-stream matching decoder across 2D-2D, 2D-3D and 3D-3D correspondence tasks in Table~\ref{table:ablation_matching_methods}. 
We compare our design to four alternative commonly adopted matching paradigms using the same small-scale model but replacing our matching  decoder with \emph{nearest neighbor}, \emph{global matching} (similar to~\cite{MASt3R,zhang2023gmsf}), \emph{regression} (similar to~\cite{CroCov2, UniFlowMatch_2025}) and \emph{sequence concatenation} (similar to COTR~\cite{COTR}). 
As shown in Table~\ref{table:ablation_matching_methods}, the regression  
and sequence concatenation methods show the worst results across all tasks. Nearest neighbor matching with dense features underperform on 2D-3D and 3D-3D tasks compared to our approach. While \emph{global matching}  achieves comparable results to our method, it is computationally expensive as it relies on large (full) feature resolution and takes approximately 2$\times$ the training duration compared to our method.


We further ablate different design choices for our model, where we \emph{progressively} add different components. The results are summarized in
Table~\ref{table:ablation_two_stream}. Starting with Setup ~\hyperref[case:I]{\RNum{1}} (baseline), the model has 8 matching decoder layers, 16 attention heads ($H=16$), vanilla attention, $D=256$, and no feature upsampling. It is trained for 30 epochs on the MegaDpeth dataset using 100 keypoint queries and 68,400 2D-2D samples per epoch.  We make the following ablations. Setup~\hyperref[case:II]{\RNum{2}} replaces vanilla attention with Gaussian attention, leading to better results. Setup~\hyperref[case:III]{\RNum{3}} increases keypoint queries per training sample from 100 to 800 to supervise the model, showing improved accuracy. Recognizing that single-head attention approximates nearest neighbor matching, Setup~\hyperref[case:IV]{\RNum{4}} tests this approach showing further performance improvements. Setup~\hyperref[case:V]{\RNum{5}} adds contrastive loss supervision, improving feature descriptor quality. Setup~\hyperref[case:VI]{\RNum{6}} upsamples spatial resolution by $4\times$ using MLP with Pixel Shuffle~\cite{PixelShuffle}, leading to great improvement in accuracy.  Finally, Setup~\hyperref[case:VI]{\RNum{7}} keeps the embedding dimension $D=256$ after upsampling. This represents our chosen configuration for the final model.

\begin{table}[t]
\small
\setlength{\tabcolsep}{3pt}
    \caption{\textbf{Image-to-Image (2D-2D) matching comparison on MegaDepth-1500 and ScanNet-1500.} 
    \textcolor{gray}{Gray text} indicates ScanNet~\cite{dai2017scannet} was part of the training datasets. \textbf{Bold} and \underline{underline} highlights best and second best results.}
    \vspace{-5mm}
    \label{tab:relative pose estimation 2d2d}
    \begin{center}
    \begin{tabular}{lccccccc}
    \toprule                                  
            & & \multicolumn{3}{c}{MegaDepth-1500}                                                  
            & \multicolumn{3}{c}{ScanNet-1500}  \\        
    \cmidrule(r){3-5}\cmidrule(r){6-8}
    Methods & AUC@ $\rightarrow$
            & \multicolumn{1}{c}{$5^\circ \uparrow$}               
            & \multicolumn{1}{c}{$10^\circ \uparrow$}     
            & \multicolumn{1}{c}{$20^\circ \uparrow$}  
            & \multicolumn{1}{c}{$5^\circ \uparrow$}  
            & \multicolumn{1}{c}{$10^\circ \uparrow$}      
            & \multicolumn{1}{c}{$20^\circ \uparrow$} \\ 
    \midrule
    
    \multicolumn{2}{l}{SP + SG~\cite{SuperPoint, SuperGlue}}                 & 49.7          & 67.1           & 80.6          & 16.2           & 32.8           &  49.7           \\
    \multicolumn{2}{l}{SP + LG~\cite{SuperPoint, LightGlue}}                 & 51.0          & 68.1           & 80.7          & 14.8           & 30.8           &  47.5           \\
    \multicolumn{2}{l}{LoFTR~\cite{LoFTR}}                         & 52.8          & 69.2           & 81.2          & 16.9           & 33.6           &  50.6           \\
    \multicolumn{2}{l}{Efficient LoFTR~\cite{Efficient_LoFTR}}     & 56.4          & 72.2           & 83.5          & 19.2           & 37.0           & 53.6           \\
    \multicolumn{2}{l}{ASpanFormer~\cite{ASpanFormer}}             & 55.3          & 71.5           & 83.1          & 19.6           & 37.7           & 54.4           \\   
    \multicolumn{2}{l}{DKM~\cite{DKM}}                             & \underline{60.4}          & \underline{74.9}           & \underline{85.1}          & 26.6           & 47.1           & 64.2           \\   
    \multicolumn{2}{l}{RoMa~\cite{RoMa}}                           & \textbf{62.6} & \textbf{76.7}  & \textbf{86.3} & {28.9}         & {50.4}         & {68.3}           \\
    \multicolumn{2}{l}{MASt3R~\cite{MASt3R}}                         & 53.1          & 70.0           &  82.4         & \textcolor{gray}{\textbf{34.1}}         & \textcolor{gray}{\textbf{57.1}}         & \textcolor{gray}{\textbf{74.3} }     \\
    \multicolumn{2}{l}{VGGT~\cite{VGGT}}                             & -             & -              & -             & \textcolor{gray}{33.9}  & \textcolor{gray}{55.2}  & \textcolor{gray}{73.4}  \\
    \multicolumn{2}{l}{UFM560-refine~\cite{UniFlowMatch_2025}}       & -          & -           & -          &  \textbf{31.6}           & \textbf{54.1}       &   70.9  \\
    \midrule					
    \multicolumn{2}{l}{\textbf{Ours (stage 1)}} & 55.5 & 71.1 & 82.8 & \underline{29.5} & \underline{52.6} &  \underline{71.2} \\
    \multicolumn{2}{l}{\textbf{Ours (stage 2)}} &  54.2  &  69.8  & 81.8 & 29.1 & 52.5 & \textbf{71.3} \\

    \bottomrule                                                
    \end{tabular}
    \end{center}
    \vspace{-6mm}
\end{table}

\begin{table}[t]
\small
    \caption{\textbf{Visual localization results (2D-2D matching) on the InLoc~\cite{InLoc} dataset}. We report the percentage of query images localized within $0.25/0.5/1.0$ meters and $2/5/10$ degrees of the ground-truth pose (higher is better). \textbf{Bold} and \underline{underline} highlights best and second best results.}
    \vspace{-4mm}
    \label{tab:visual localization 2d2d}
    \begin{center}
    \begin{tabular}{lcc} 
    \toprule                                                    
                  Method 
                  & DUC1
                  & DUC2   \\ 
    \midrule
    {SP + SG~\cite{SuperPoint, SuperGlue}}     & 49.0 / 68.7 / 80.8                      & 53.4 / 77.1 / 82.4  \\
    {LoFTR~\cite{LoFTR}}                       & 47.5 / 72.2 / 84.8                      & 54.2 / 74.8 / 85.5  \\
    {PATS~\cite{PATS}}                         & 55.6 / 71.2 / 81.0                      & 58.8 / 80.9 / 85.5  \\
    {DKM~\cite{DKM}}                           & 51.5 / 75.3 / 86.9                      & 63.4 / 82.4/ 87.8   \\
    {CasMTR{~\cite{CasMTR}}}                   & 53.5 / 76.8 / 85.4                      & 51.9 / 70.2 / 83.2  \\
    {RoMa~\cite{RoMa}}                         & \textbf{60.1} / \textbf{79.3} / \underline{89.9}    & \underline{66.4} / \underline{83.2} / \underline{87.8}  \\
    {MASt3R~\cite{MASt3R}}                     & \underline{56.1} / \textbf{79.3} / \textbf{90.9}    & \textbf{71.0} / \textbf{87.0} / \textbf{91.6}  \\
    \midrule
    \textbf{Ours (stage 2)}                    & \underline{56.1} / \textbf{79.3} / 89.4              &  61.1 / 80.2 / 84.0  \\
    \bottomrule
    \end{tabular}
    \end{center}
    \vspace{-8mm}
\end{table}

We further investigate the small-scale model from Setup~\hyperref[case:VI]{\RNum{7}} with different number of decoder layers and feature upsampling ratios. 
As can be seen in Fig.~\ref{fig:ablation plots}, our matching decoder benefits from stacking multiple dual-stream Transformer layers, \emph{which validates our core contribution.}
The performance plateaus after 3 decoder layers likely because of the limited capacity in the small-scale model.
Regarding the feature upsampling ratio, we can see that larger resolutions are generally helpful until $8\times$ upsampling. To balance the accuracy and efficiency, we use $4\times$ upsampling.

\subsection{Comparisons with Other Methods}
\label{subsec: benchmarks}
In this section, we compare our large-scale unified model against other task-specific methods.

\noindent\textbf{2D-2D Benchmarks.}
We compare our model  with recent 2D-2D SOTA approaches~\cite{RoMa, MASt3R, VGGT, UniFlowMatch_2025} 
on two-view geometry benchmarks MegaDepth-1500~\cite{megadepth} and ScanNet-1500~\cite{dai2017scannet} in 
Table~\ref{tab:relative pose estimation 2d2d}.
We use valid keypoints detected by RoMa~\cite{RoMa} to query our model.
Our unified model shows strong generalization on ScanNet-1500 achieving an AUC$@ 20 ^\circ$ score of over 71
among methods which are not supervised on the ScanNet dataset itself. Our method outperforms MASt3R~\cite{MASt3R}
using the same coarse-to-fine inference pipeline on MegaDepth-1500 and it is the third best model compared
to other task-specific SOTA 2D-2D models. It is important to note that DKM~\cite{DKM} and RoMa~\cite{RoMa} achieve better results on the MegaDepth benchmark by warping high-resolution image features for improved sub-pixel accuracy. However, this approach is inapplicable to 2D-3D correspondence, as warping from a 2D grid to 3D is undefined.
For the InLoc~\cite{InLoc} benchmark, we query our model with keypoints sampled from an uniformly spaced grid. Our model achieves competitive results to SOTA models as reported in Table~\ref{tab:visual localization 2d2d}.

\begin{table}[t]
\small
\setlength{\tabcolsep}{3pt}
    \caption{\textbf{Image-to-Point (2D-3D) matching comparison on 7Scenes and RGB-D Scenes V2.} 
    We report Inlier Ratio (IR), Feature Matching Ratio (FMR) and the Registration Recall (RR). \textbf{Bold} and \underline{underline} highlights best and second best results.}
    \vspace{-2mm}
    \label{tab:image-to-point benchmarks}
    \centering
    \begin{tabular}{lccccccc}
    \toprule
            & & \multicolumn{3}{c}{7Scenes}                                                  
            & \multicolumn{3}{c}{RGB-D Scenes V2}  \\
    \cmidrule(r){3-5}\cmidrule(r){6-8}
    Methods &
            & \multicolumn{1}{c}{IR $\uparrow$}               
            & \multicolumn{1}{c}{FMR $\uparrow$} 
            & \multicolumn{1}{c}{RR $\uparrow$}               
            & \multicolumn{1}{c}{IR $\uparrow$}               
            & \multicolumn{1}{c}{FMR $\uparrow$} 
            & \multicolumn{1}{c}{RR $\uparrow$} \\
    \midrule
        \multicolumn{2}{l}{FCGF-2D3D~\cite{FCGF}}          &   22.8          & 78.8          & 61.4          &  8.1           & 27.1          &  30.4            \\
        \multicolumn{2}{l}{Predator-2D3D~\cite{Predator}}  &   23.4          & 77.5          & 48.5          &  15.7          & 59.6          &  38.4            \\
        \multicolumn{2}{l}{P2-Net~\cite{P2-net}}           &   31.7          & 79.0          & 65.7          &  12.2          & 65.9          &  30.2            \\
        \multicolumn{2}{l}{2D3D-MATR~\cite{2D3D-MATR}}     &   50.1          & 92.1          & 75.8          &  32.4          & 90.8          &  56.4            \\
        \multicolumn{2}{l}{B2-3Dnet~\cite{Bridge-2D3D}}    &   50.9          & \textbf{93.1} & 77.7 &  \underline{35.1}        &  \underline{94.4}     &  63.4            \\
        \multicolumn{2}{l}{FreeReg~\cite{FreeReg}}         &       -         &       -        &      -      &   30.9         & 82.0           &  57.3                  \\
        \multicolumn{2}{l}{Diff-Reg~\cite{diffregv2}}   &  \underline{59.1}  &   92.5     &    \underline{83.8}      &    -            & -              &  \underline{87.4}     \\
    \midrule

    \multicolumn{2}{l}{\textbf{Ours (stage 2)}} & \textbf{82.4} & \underline{93.0} & \textbf{91.0} & \textbf{83.6} & \textbf{97.0} & \textbf{92.5} \\

    \bottomrule
    \end{tabular}
    \vspace{-5mm}
\end{table}

\begin{figure*}[t]
\centering
\begin{minipage}{0.63\textwidth}
    \centering
    \footnotesize
    \setlength{\tabcolsep}{1.5pt}
    \captionof{table}{\textbf{Point-to-Point (3D-3D) matching comparison on  3DMatch, 3DLoMatch, and ModelNet.} We report Inlier Ratio (IR), Feature Matching Ratio (FMR), Registration Recall (RR), Relative Rotation Error (RRE), Relative Translation Error (RTE) and Chamfer Distance (CD). \textbf{Bold} and \underline{underline} highlights best and second best results.}
    \label{tab:point-to-point benchmarks}
    \begin{tabular}{lccccccccccc}
    \toprule
            & & \multicolumn{3}{c}{3DMatch}                                                  
            & \multicolumn{5}{c}{3DLoMatch}  
            & \multicolumn{2}{c}{ModelNet}  \\
    \cmidrule(r){3-5}\cmidrule(r){6-10} \cmidrule(r){11-12}
    Methods &
            & \multicolumn{1}{c}{IR $\uparrow$}               
            & \multicolumn{1}{c}{FMR $\uparrow$} 
            & \multicolumn{1}{c}{RR $\uparrow$}
            & \multicolumn{1}{c}{IR $\uparrow$}               
            & \multicolumn{1}{c}{FMR $\uparrow$} 
            & \multicolumn{1}{c}{RR $\uparrow$} 
            & \multicolumn{1}{c}{RRE[$^\circ$] $\downarrow$}
            & \multicolumn{1}{c}{RTE[$m$] $\downarrow$} 
            & \multicolumn{1}{c}{RRE $\downarrow$}
            & \multicolumn{1}{c}{RTE $\downarrow$} \\
    \midrule
        \multicolumn{2}{l}{FCGF~\cite{FCGF}} & 56.8 & 97.4 & 85.1 & 21.4 & 76.6 & 40.1 & 3.147 & 0.100 & - & - \\
        \multicolumn{2}{l}{D3Feat~\cite{bai2020d3feat}} & 39.0 & 95.6 & 81.6 & 13.2 & 67.3 & 37.2 & 3.361 & 0.103 & - & - \\
        \multicolumn{2}{l}{CofiNet~\cite{yu2021cofinet}} & 49.8 & 98.1 & 89.3 & 24.4 & 83.1 & 67.5 & - & - & - & - \\
        \multicolumn{2}{l}{Predator~\cite{Predator}} & 58.8 & 96.6 & 89.0 & 26.7 & 78.6 & 59.8 & 3.048 & 0.093 & 1.739 & 0.019 \\
        \multicolumn{2}{l}{RegTR~\cite{Regtr}} & - & - & 92.0 & - & - & 64.8 & 2.827 & \underline{0.077} & 1.473 & 0.014 \\
        \multicolumn{2}{l}{GeoT~\cite{GeoTransformer}} & 70.3 & 97.7 & 91.5 & 43.3 & 88.1 & 74.0 & \underline{2.547} & \textbf{0.074} & 1.568 & 0.018 \\
        \multicolumn{2}{l}{RoITr~\cite{roltr}} & 82.6 & 98.0 & 91.9 & 54.3 & 89.6 & 74.8 & - & - & - & - \\
        \multicolumn{2}{l}{PEAL-3D~\cite{peal3d}} & 73.3 & \textbf{99.0} & 94.6 & 49.0 & 87.6 & 79.0 & 2.788 & 0.087 & - & - \\
        \multicolumn{2}{l}{Diff-Reg~\cite{diffregv2}} & 30.9 & 96.3 & 95.0 & 9.6 & 69.6 & 73.8 & - & - & - & - \\
    \midrule
    \multicolumn{2}{l}{\textbf{Ours (stage 1)}}  & \underline{95.1} & \underline{98.5} & \underline{97.4} & \textbf{77.3} 
                                                        & \textbf{93.2} & \textbf{86.7} & 3.193 & 0.168 & \textbf{1.189} & \textbf{0.011} \\
    \multicolumn{2}{l}{\textbf{Ours (stage 2)}}  & \textbf{95.2} & 98.4 & \textbf{97.5} & \underline{75.0} 
                                                            & \underline{90.8} & \underline{83.2} & \textbf{2.393} & 0.152 & \underline{1.324} & \underline{0.012} \\ 
    \bottomrule
    \end{tabular}
\end{minipage}%
\hfill
\begin{minipage}{0.36\textwidth}
    \centering
    \includegraphics[width=0.85\linewidth, trim=0 20 0 0, clip]{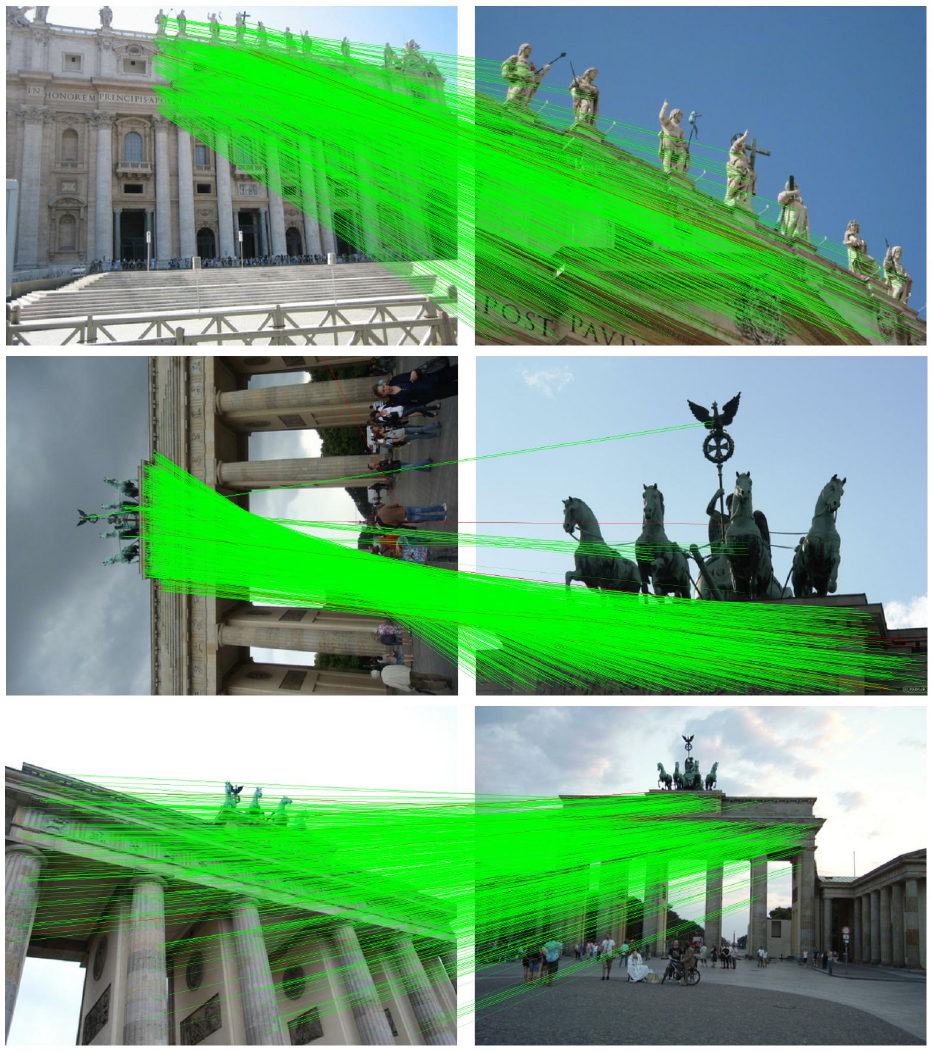}
    \vspace{-4pt}
    \captionof{figure}{\textbf{Visual results of 2D-2D matching on MegaDepth.} \textcolor{green}{Green}/\textcolor{red}{red} lines indicate inlier/outlier correspondences. Zoom in for details.}
\end{minipage}
\vspace{-3mm}
\end{figure*}

\noindent\textbf{2D-3D Benchmarks.}
For 2D-3D, we compare our unified model with SOTA on the 7Scenes~\cite{7Scenes} and 
RGB-D Scenes V2~\cite{RGB-DScenesV2} test split containing 2304 and 497 image-to-point
pairs, respectively. We use the ground-truth keypoints to query our model. 
Compared with other \emph{dataset-specific} 2D-3D methods reported in Table~\ref{tab:image-to-point benchmarks}, which are trained and evaluated separately on each dataset, 
our unified model achieves the best results, outperforming other methods by $8\%$ on \emph{registration recall} (RR) on the 7Scenes dataset.

\noindent\textbf{3D-3D Benchmarks.}
We report 3D-3D registration results on 3DMatch~\cite{3dmatch}, 3DLoMatch~\cite{3dmatch} and ModelNet~\cite{modelnet}
test split containing 1623, 1781, and 1266 point-to-point pairs, respectively. 
We use ground truth keypoints from 3DMatch for correspondence evaluation, 
and on ModelNet we query the entire source point cloud and apply a cycle consistency check with matching threshold $\tau_{\text{cycle}} = 0.02$ to filter out the set of invalid correspondences.
Similar to 2D-3D, our unified model shows a $10\%$ improvement on \emph{registration recall} (RR) in the challenging low overlap 3DLoMatch benchmark compared the best \emph{dataset-specific}
model in Table~\ref{tab:point-to-point benchmarks}. 

It is worth noting that using ground-truth keypoints for 2D-3D and 3D-3D matching does not necessarily put our model in a more advantageous position than others. 
On the one hand, existing work~\cite{Predator, 2D3D-MATR, GeoTransformer, Diff-Reg} uses ground truth transformation to align their predictions in order to evaluate their matched pairs whereas our model directly regresses the estimated correspondences in the target coordinate space.
On the other hand, we show that on ModelNet, using only cycle consistency check without relying on ground-truth keypoints leads to lower errors than other methods.

\begin{figure}[t]
  \centering
   \includegraphics[width=0.95\linewidth, trim=0 20 0 0, clip]{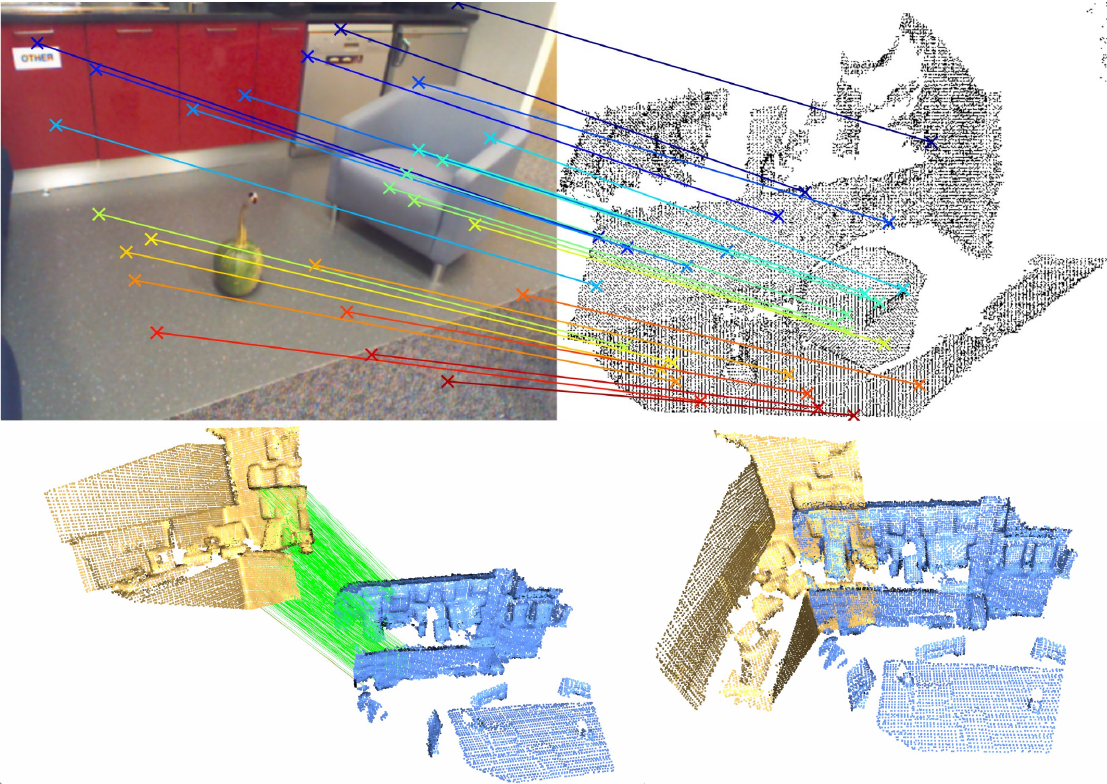}
   \vspace{-2pt}
  \caption{\textbf{Visual results of 2D-3D matching on 7Scenes (top) and 3D-3D matching on 3DLoMatch (bottom)}. On the bottom left are point cloud pairs with predicted correspondences, and on the bottom right are registered point clouds using transformations estimated via RANSAC. Zoom in for details.}
\label{fig:3d3d_visual_results}
\vspace{-8pt}
\end{figure}

\subsection{On the Joint Training of Different Tasks}

\begin{table}[b]
\scriptsize
    \vspace{-16pt}
    \caption{Single task vs. joint training performances.}
    \setlength{\tabcolsep}{1.5pt}
    \vspace{-5pt}
    \label{tab:single-vs-joint tradeoff}
    \centering
    \begin{tabular}{lcccc}
    \toprule
            & & MegaDepth (2D-2D)                                 
            & 7Scenes (2D-3D) 
            & 3DLoMatch (3D-3D)   \\
    \cmidrule(r){3-3}\cmidrule(r){4-4} \cmidrule(r){5-5}
    Methods &
            & \multicolumn{1}{c}{AUC@5$^\circ$ $\uparrow$}
            & \multicolumn{1}{c}{RR $\uparrow$}
            & \multicolumn{1}{c}{RR $\uparrow$} \\
    \midrule
    \multicolumn{2}{l}{Ours (single task)}  & \textbf{56.5} & 67.7 & 81.8 \\ 
    \multicolumn{2}{l}{Ours (joint training)}      & 54.2    & \textbf{91.0}  & \textbf{83.2}   \\ 
    \bottomrule
    \end{tabular}
    \vspace{-10pt}
\end{table}

An important question for unified models is whether joint training across tasks 
provides synergistic benefits. We compare our stage 1 and stage 2 models on 
2D-2D and 3D-3D tasks in Tables~\ref{tab:relative pose estimation 2d2d} 
and~\ref{tab:point-to-point benchmarks}. Results show that joint training on 
all three tasks (stage 2) does not consistently outperform stage 1 (2D-2D and 
3D-3D only). To investigate this, we analyzed gradient conflicts using the GCD 
metric~\cite{chai2024gcd}. While most parameters show aligned-to-orthogonal 
gradients (indicating minimal interference), normalization layers exhibit 
substantial conflicts.
This suggests that normalization layers struggle to 
accommodate the different statistical properties of 2D image and 3D 
point features when computing shared statistics across modalities. 

Despite these conflicts, our model shows significant improvement on 7Scenes (2D–3D) with joint training, as shown in Table~\ref{tab:single-vs-joint tradeoff}, indicating mutual benefits from the data-rich 2D–2D domain and demonstrating that the unified architecture provides a reasonable trade-off.
Future work could explore better normalization strategies or improved cross-modality 
alignment designs.

\section{Conclusion}
We presented UniCorrn, the first correspondence model with shared weights that unifies geometric matching across 2D-2D, 2D-3D, and 3D-3D modalities. 
Our dual-stream Transformer decoder, which decouples appearance and positional features, enables robust correspondence learning across heterogeneous representations. 
Trained jointly on diverse data, UniCorrn achieves competitive 2D-2D performance and sets new state-of-the-art on 2D-3D and 3D-3D matching tasks.
This work demonstrates the feasibility and 
benefits of unified correspondence modeling.
We believe this work represents an important step toward general-purpose 
correspondence models and hope it inspires further research in unified 
geometric understanding across different modalities.


\section{Acknowledgment}
\label{sec:acknowledge}
This project was partially supported by the National Science Foundation under Award IIS-2310254.

{
    \small
    \bibliographystyle{ieeenat_fullname}
    \bibliography{main}
}

\clearpage

\clearpage
\setcounter{page}{1}
\maketitlesupplementary

\appendix

\begin{figure}[t]
  \centering
   \includegraphics[width=\linewidth, trim=30 0 0 5, clip]{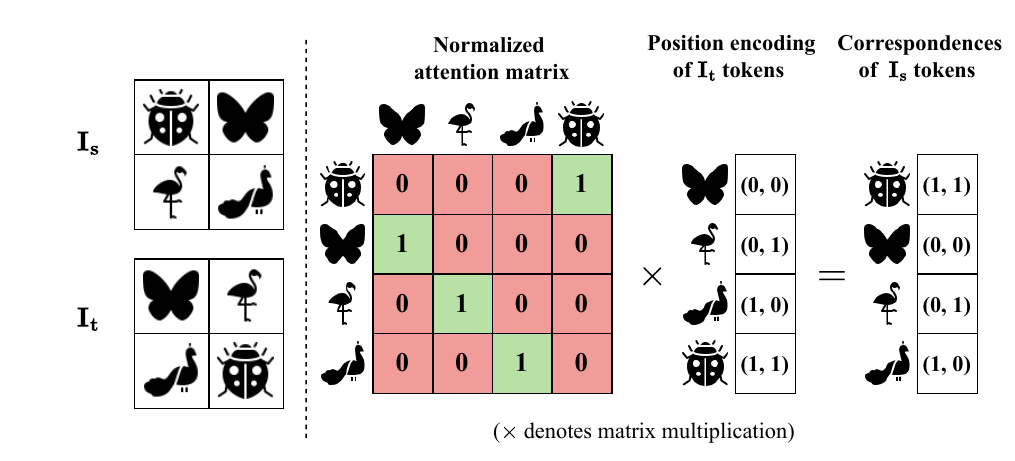}
   \vspace{-2pt}
  \caption{\textbf{Illustration of estimating correspondence with attention}. Here each animal symbol denotes a pixel (so both $\Is$ and $\It$ have $2\times 2$ pixels.).}
\label{fig:attention illustration}
\vspace{-5pt}
\end{figure}

\section{Attention as a Learnable Matching Cost}
In figure~\ref{fig:attention illustration}, we show an illustration of using attention to estimate correspondences with a toy example.
Let's consider two input images $\Is$ and $\It$.  
The attention map $\vA$ between them is computed as the \texttt{Softmax}-normalized dot product of the flattened inputs.
The attention map is row-normalized and one-hot in each row in an ideal case, where the position of 1 corresponds to the correct matching pixel. 
If we set the vector $\vV$ in Transformer to the \emph{absolute positional encoding} of every pixel in $\It$, as shown in Fig.~\ref{fig:attention illustration}, the output $\vA\vV$ contains the positional encoding of the correct corresponding pixels in $\It$ for every pixel in $\Is$.

The attention matrix is similar to the normalized version of the learnable cost volume studied in~\cite{xiao20learnable}.
In practice, while features may not be perfectly discriminative as demonstrated in this example, the methodology of using attention matrix as a matching cost function still applies.



\section{Further Details on Matching Decoder}

\subsection{Gaussian Attention}
In the paper, we propose Gaussian attention 
in replace of vanilla attention~\cite{vaswani2017attention} within our matching decoder. The attention logits are computed using pairwise squared $\texttt{L2}$ distance is formulated as:
\begin{equation}
    a_{ij} = -\frac{\|Q_i - K_j\|^2}{D},
\end{equation}
where $Q$ and $K$ are query and key tokens and $D$ is the embedding dimension. 
Furthermore, if we took \texttt{Softmax} function into consideration to get the normalized attention scores, the equation becomes:
\begin{equation}
    \mathbf{A}_{ij} = \frac{\text{exp}(a_{ij})}{\sum_k\text{exp}(a_{ik})}.
\end{equation}
Here, $\text{exp}(a)$ fits into the general formulation of the Gaussian kernel.

\subsection{InfoNCE Loss}
We provide further details of computing InfoNCE~\cite{infoNCE} loss. 
For a given pair of source and target feature descriptors $\fs^{desc}$ and $\ft^{desc}$, respectively, the InfoNCE loss over the set of ground-truth correspondences 
$\mathcal{M} = \{{\bar\ks(i), \bar\kt(i)}\}^N_{i=1}$ is given by:


\begin{equation}
\begin{aligned}
    \mathcal{L}_{c}(\fs^{desc}, \ft^{desc}) &= -\sum^N_{i=1} \text{log}\frac{d(\bar\ks(i), \bar\kt(i))}{\sum^N_{j=1} d(\bar\ks(j), \bar\kt(i))} \\
    &+ \text{log}\frac{d(\bar\ks(i), \bar\kt(i))}{\sum^N_{j=1} d(\bar\ks(i), \bar\kt(j))},
\end{aligned}
\end{equation}

\begin{equation}
    \text{with } d(\bar\ks, \bar\kt) = \tau^{-1}||\fs^{desc}(\mathbf{\bar\ks}) - \ft^{desc}(\bar\kt)||_2,\notag
\end{equation}
where $\tau$ is a temperature hyperparameter. Similarly, we compute the InfoNCE loss for $\mathcal{L}_{c}(\fk, \ft^{desc})$.


\subsection{Pseudo Point Cloud Data}

In Table~\ref{table:ablation pseudo pcd}, we show the effectiveness of using pseudo point cloud data for the 2D3D and 3D3D tasks. The pseudo point cloud is generated from dense depth maps, where depth is projected to dense 3D points and sampled with equal strides to resemble the sparse structure of the 3D benchmark datasets. As our approach is data-driven, jointly training with pseudo-point cloud data enables our model to reach SOTA performance.

\begin{table}[h]
\small
\setlength{\tabcolsep}{3pt}
\centering
\renewcommand{\midrule}{\noalign{\vskip 3pt\hrule height 0.8pt\vskip 3pt}}
\caption{\textbf{Effectiveness of pseudo point cloud data} for 2D-3D and 3D-3D task. The pseudo data is sampled from ScanNet++~\cite{scannetpp} depth maps.}
\label{table:ablation pseudo pcd}
\begin{tabular}{c|ccc|ccc} 
\toprule
\multirow{2}{*}{}      \multirow{2}{*}{\begin{tabular}[c]{@{}l@{}}Pseudo \\ Point Cloud  \end{tabular}}
                       & \multicolumn{3}{c|}{7Scenes (2D-3D)} 
                       & \multicolumn{3}{c}{3DLoMatch (3D-3D)} \\ 
                       
\cmidrule{2-4}\cmidrule{5-7}
                       & IR $\uparrow$  
                       & FMR $\uparrow$ 
                       & RR  $\uparrow$ 
                       & IR  $\uparrow$
                       & FMR $\uparrow$ 
                       & RR  $\uparrow$ \\ 
\midrule

\xmark      & 12.9 & 49.5 & 15.4 &  51.1 & 83.2 & 73.2 \\ 	
\cmark                 & \textbf{66.3}	&   \textbf{88.2}	&   \textbf{77.8}  & \textbf{70.5}	&   \textbf{90.1}	&  \textbf{81.8}   \\
\bottomrule
\end{tabular}
\end{table}

\subsection{Auxiliary Supervision}

\begin{figure*}[t]
\vspace{-6mm}
  \centering
    \begin{overpic}[width=0.7\linewidth, trim=0 20 0 0, clip]{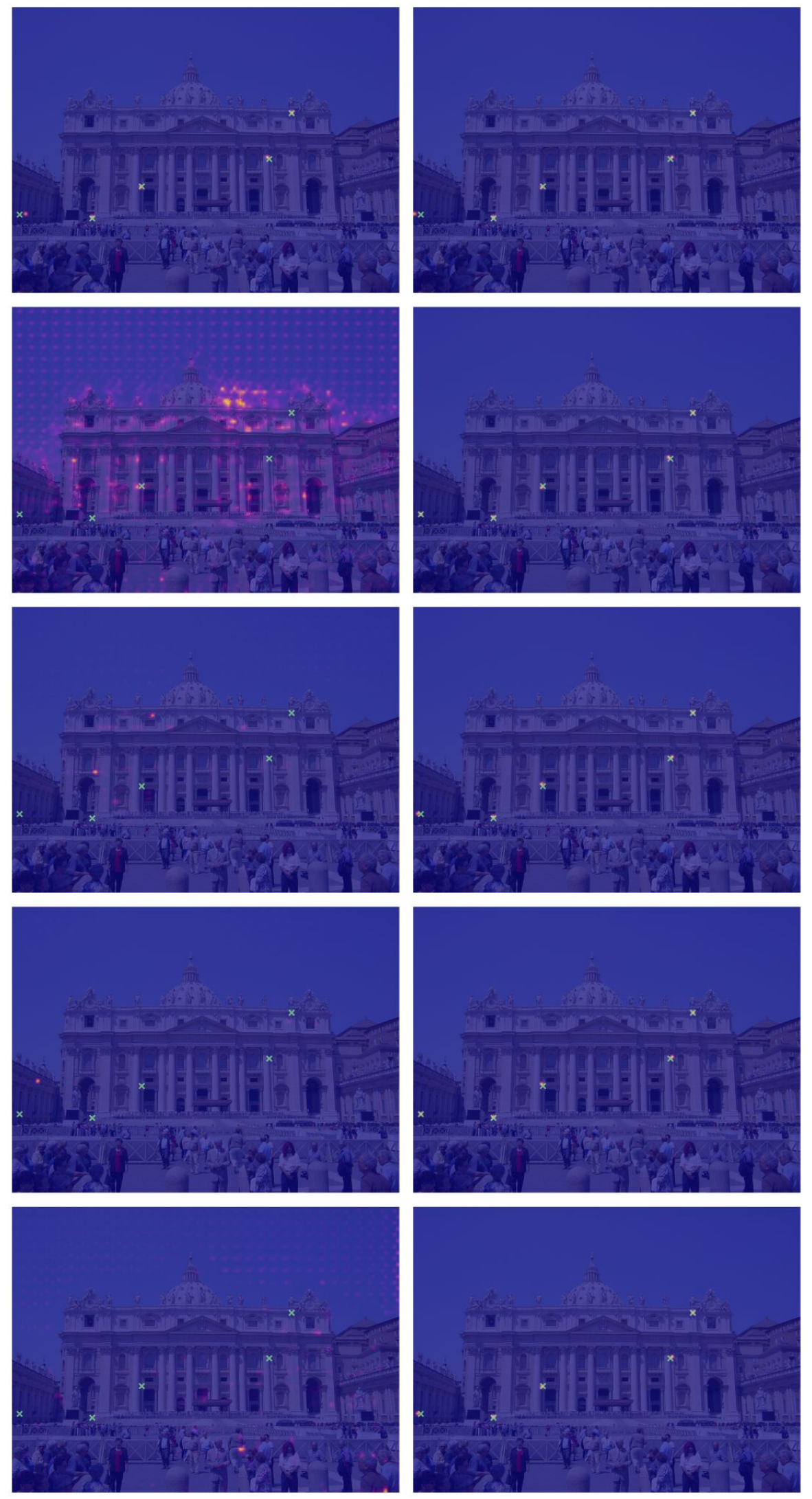}
        \put(-6, 89){\textbf{Layer 1}}
        \put(-6, 69){\textbf{Layer 2}}
        \put(-6, 48){\textbf{Layer 3}}
        \put(-6, 28){\textbf{Layer 4}}
        \put(-6, 8){\textbf{Layer 5}}
        \put(4,-2){\textbf{Without auxiliary supervision}}
        \put(32,-2){\textbf{With auxiliary supervision}}
    \end{overpic}
    \vspace{4mm}
  \caption{\textbf{Per-layer attention heatmap comparison for the effectiveness of auxiliary supervision}.
  \textcolor{green}{Green} markers indicates the model's predicted coordinates.
  Zoom in for more details.
  }
\label{fig:attn_map_full}
\end{figure*}

In our training objective, we use intermediate predictions by applying the attention matrix directly over the target coordinates for auxiliary supervision.
As shown in Tab.~\ref{table:ablation_gm_aux}, the auxiliary loss produced substantial performance improvement with a single matching decoder layer and also improved the results while scaling up the number of layers.
In Figure.~\ref{fig:attn_map_full} we visualize the attention heatmaps for each decoder layer along with the final predicted coordinates from the model. 
The heatmaps show a clear difference: without auxiliary supervision, attention patterns are random across layers, while with auxiliary supervision, query tokens consistently attend to their corresponding predicted coordinates. This shows how the dual-stream attention
propagates through the matching decoder layers.

\begin{table}[ht]
\small
\setlength{\tabcolsep}{4pt}
\centering
\renewcommand{\midrule}{\noalign{\vskip 3pt\hrule height 0.8pt\vskip 3pt}}

\caption{\textbf{Effectiveness of auxiliary loss $\mathcal{L}_{aux}$}.} 
\label{table:ablation_gm_aux}
\begin{tabular}{lc|ccc} 
\toprule
\multirow{2}{*}{\begin{tabular}[c]{@{}l@{}}Number of \\Layers \end{tabular}} 
& & \multicolumn{3}{c}{MegaDepth-1500}  \\
\cmidrule{3-5}
& $\mathcal{L}_{aux}$ & $5^\circ \uparrow$ & $10^\circ \uparrow$ & $20^\circ \uparrow$ \\
\midrule 
1 & \xmark & 28.8 & 45.3 & 61.3 \\
1 & \cmark & 47.7 & 64.2 & 77.2 \\
5 & \xmark & 48.5 & 65.1 & 77.9 \\
5 & \cmark & 50.6 & 67.1 & 79.6 \\
\bottomrule
\end{tabular}
\end{table}

\subsection{Additional details on model and training}

We train two models with two different capacities.
For the small-scale model, we employ 12-layer ViT~\cite{ViT} and PTv3~\cite{PTv3} transformers as image and point cloud backbones, respectively, along with an 8-layer shared Transformer for feature fusion encoder.
We ablate various configurations of our matching transformer decoder using this setup in Section~\ref{subsec: ablation study}. 
The large-scale model extends these architectures to 24 and 14 layers for the ViT and PTv3 backbones, and 12 and 8 layers for the feature fusion encoder and matching decoder, respectively. We train the large-scale unified model (600M parameters) in two stages. In the first stage, the model
is initialized with the pre-trained weights of CroCo v2~\cite{CroCov2} and jointly trained on 2D-2D and 3D-3D tasks with the AdamW optimizer for 40 epochs. This stages uses $384,000$ 2D-2D pairs and $384,000$ of 3D-3D pairs. The second stage is trained
on all three tasks for 30 epochs with $60,000$ samples per task per epoch. The input images are resized to $512 \times 384$ for the 2D-2D and 2D-3D tasks.
The training runs on $8\times$H100 GPUs with stage 1 taking 7 days and stage 2 taking 4 days.

\noindent In Table~\ref{table:suppl_model_details}, we provide the configurations of each module
for our small and large-scale models. Table~\ref{table:suppl_training_hparams} contains
the hyperparameters used for the two stage large-scale training. Finally, Table~\ref{table:suppl_dataset}
shows the mixture of 2D-2D, 2D-3D and 3D-3D datasets along with the pseudo data samples used in
each stage of large-scale training. We further oversample the 2D-3D and 3D-3D pairs to match the total number
of pairs used for the 2D-2D task so that the model can be jointly trained.

\section{Generalization to unseen correspondence tasks} 
Our model may generalize to other geometry matching tasks, like optical flow without any fine-tuning. On the Sintel final training split, our model achieves an end-point error (EPE) of 5.2 with zero-shot inference (specialist model RAFT reports EPE of 2.71).
This is significant because our model was trained exclusively on static, photorealistic imagery, making Sintel's dynamic motion and stylized rendering strictly out-of-distribution.
For other correspondence task, like semantic matching, fine-tuning is required.
In fact, unifying both geometric and semantic understanding with a single model by training on all different data is an exciting direction to go.

\section{Inference time and memory usage}
The memory footprint of our unified model is $\sim2.6G$ which is $3.5\times$ less than the combined memory usage of the specialized models.
We report the inference time comparisons with specialized models in the table below, measured on an RTX A5000.

\begin{table}[h]
\footnotesize
    \setlength{\tabcolsep}{2pt}
    \caption{Inference time in milliseconds(ms) on RTX A5000. Our method uses 5000 keypoint queries. Diff-Reg~\cite{diffregv2} uses existing models  for 2D-3D~\cite{2D3D-MATR} and 3D-3D~\cite{lepard} feature descriptors.}
    \label{tab:compute resources}
    \centering
    \begin{tabular}{lllll}
    \toprule
            & & ScanNet (2D-2D)                                 
            & 7Scenes (2D-3D)
            & 3DMatch (3D-3D)  \\
    \midrule
    \multicolumn{2}{l}{Ours }         & 329 ms          &  390 ms &  320 ms \\ 
    \multicolumn{2}{l}{Specialized}   & 203 ms (RoMa)   &  1140 ms (Diff-Reg)  &  603 ms (Diff-Reg)  \\
    \bottomrule
    \end{tabular}
\end{table}

\section{Additional visual results}
\begin{figure*}[t]
\centering
\begin{minipage}[t]{0.33\textwidth}
\centering
\includegraphics[width=\linewidth]{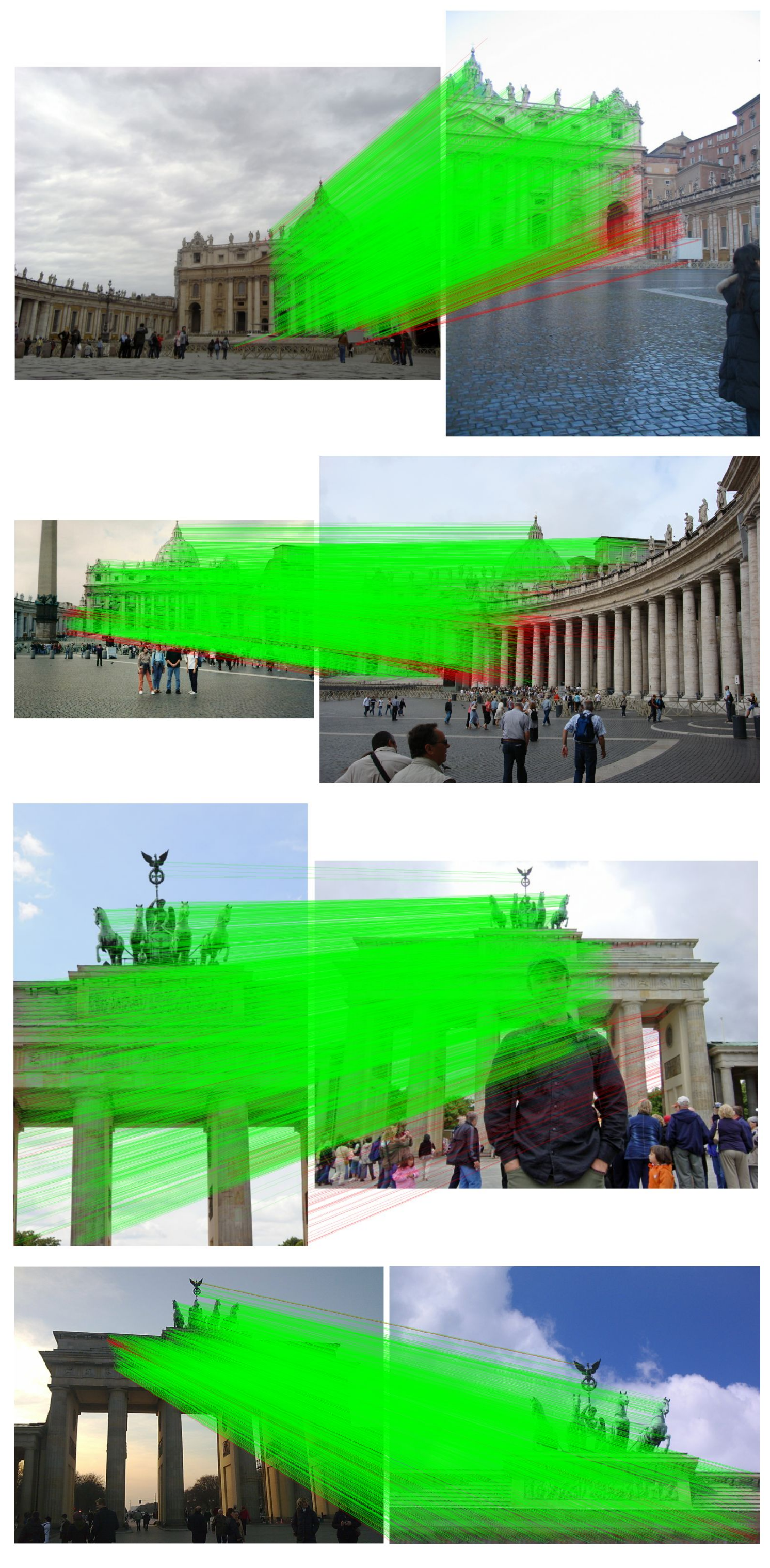}
RoMA~\cite{RoMa}
\end{minipage}
\begin{minipage}[t]{0.33\textwidth}
\centering
\includegraphics[width=\linewidth]{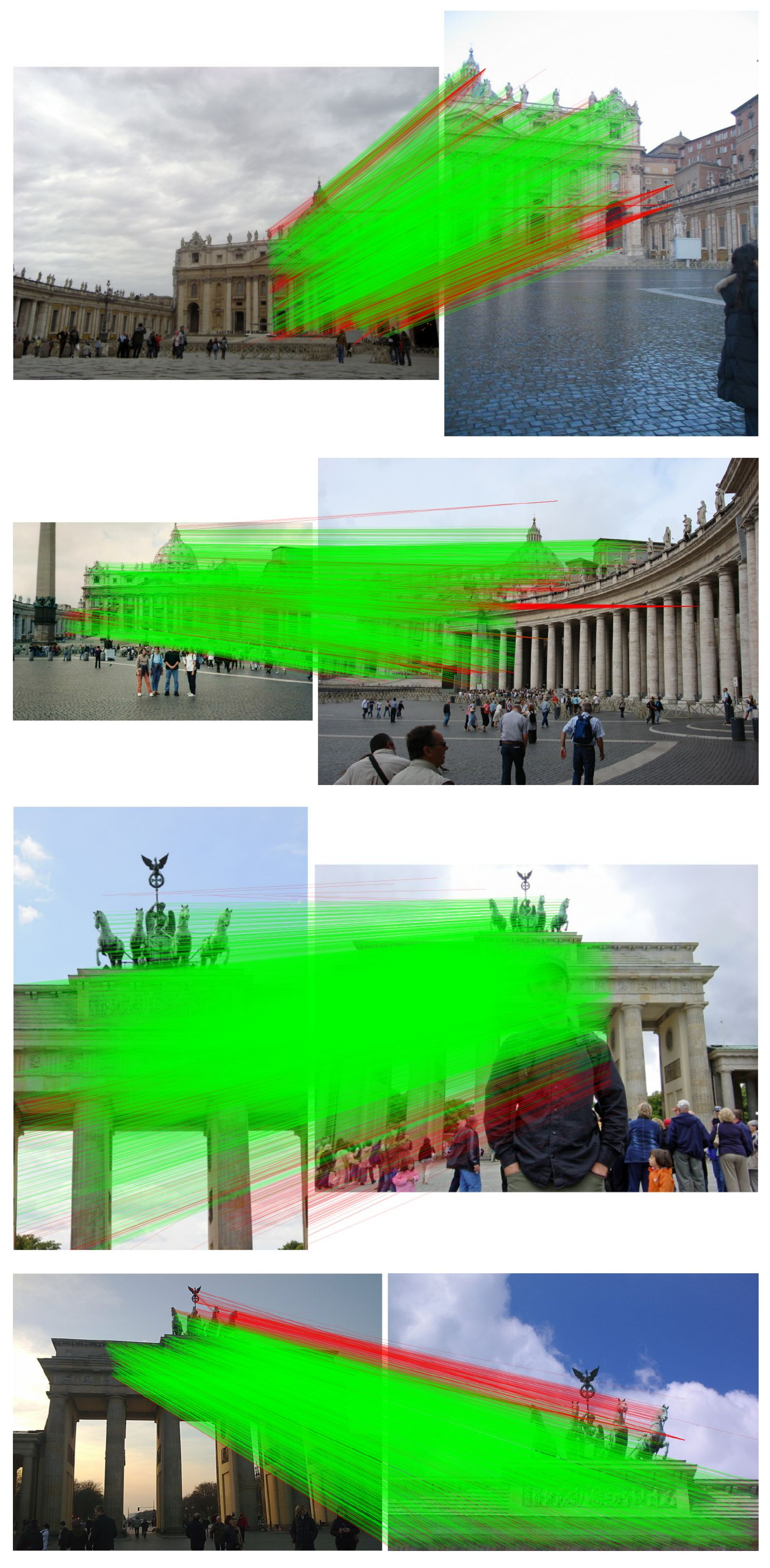}
MASt3R~\cite{MASt3R}
\end{minipage}
\begin{minipage}[t]{0.33\textwidth}
\centering
\includegraphics[width=\linewidth]{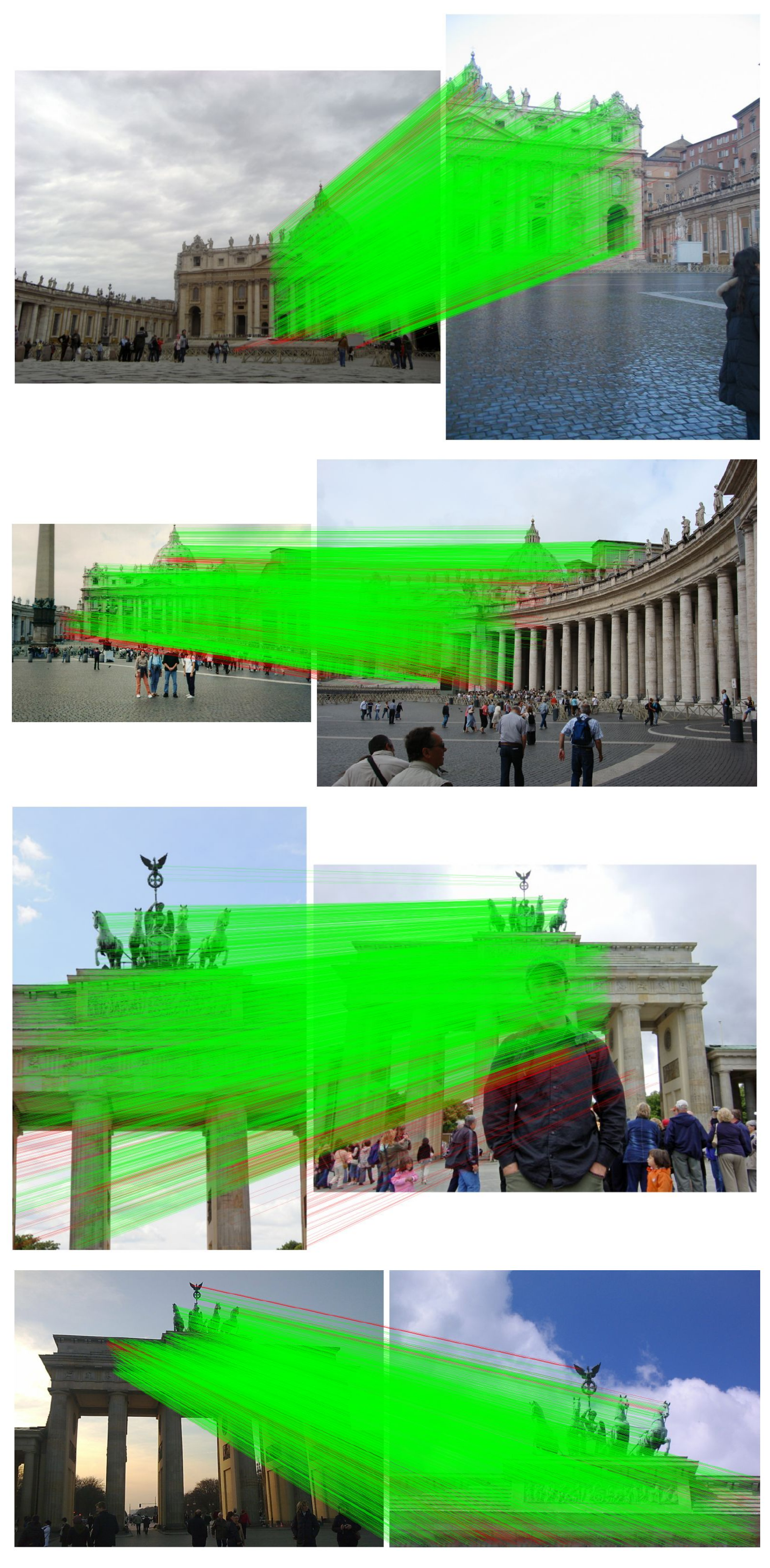}
\textbf{UniCorrn (Ours)}
\end{minipage}
\caption{\textbf{2D-2D qualitative comparisons on the MegaDepth-1500 benchmark.}
\textcolor{green}{Green} and \textcolor{red}{red} lines indicate accepted and rejected correspondences by the RANSAC essential matrix estimation, respectively. 
Zoom in for details.}
\label{fig: 2d2d qualitative comparison}
\end{figure*}

\begin{figure}[h]
  \centering
   \includegraphics[width=0.95\linewidth, trim=0 0 0 0, clip]{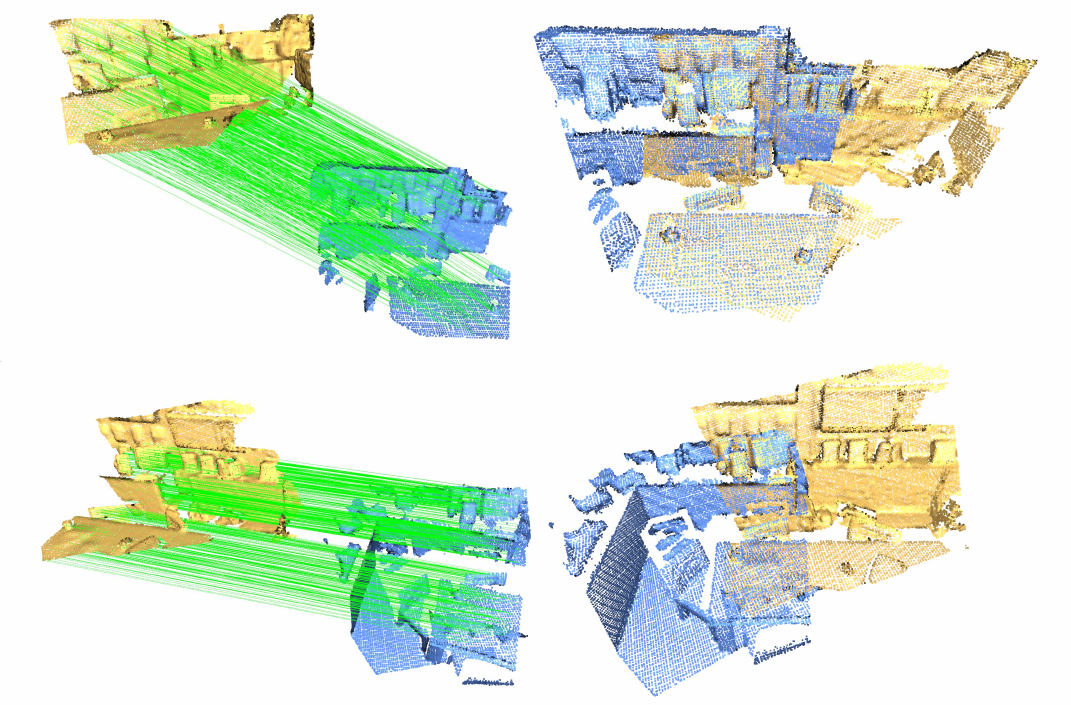}
  \caption{\textbf{Visual results of 3D-3D matching on 3DMatch (top) and 3DLoMatch (bottom)}. On the left are point cloud pairs with predicted correspondences, and on the right are registered point clouds using transformations estimated via RANSAC.}
\label{fig:3d3d_additional_visual_results}
\end{figure}

We show qualitative comparison with state-of-the-art 2D-2D matching methods RoMa~\cite{RoMa} and MASt3R~\cite{MASt3R} in Figure~\ref{fig: 2d2d qualitative comparison}. Figure~\ref{fig:InLoc_qualitative_results} shows the correspondences for different confidence thresholds on two examples from the InLoc~\cite{InLoc} benchmark. Additionally, we provide visual results for 2D-3D and 3D-3D in Figure~\ref{fig:2d3d_visual_results} and Figure~\ref{fig:3d3d_additional_visual_results}, respectively.

\twocolumn[{
    \renewcommand\twocolumn[1][]{#1}%
  \centering
   \includegraphics[width=0.65\linewidth, trim=0 0 0 0, clip]{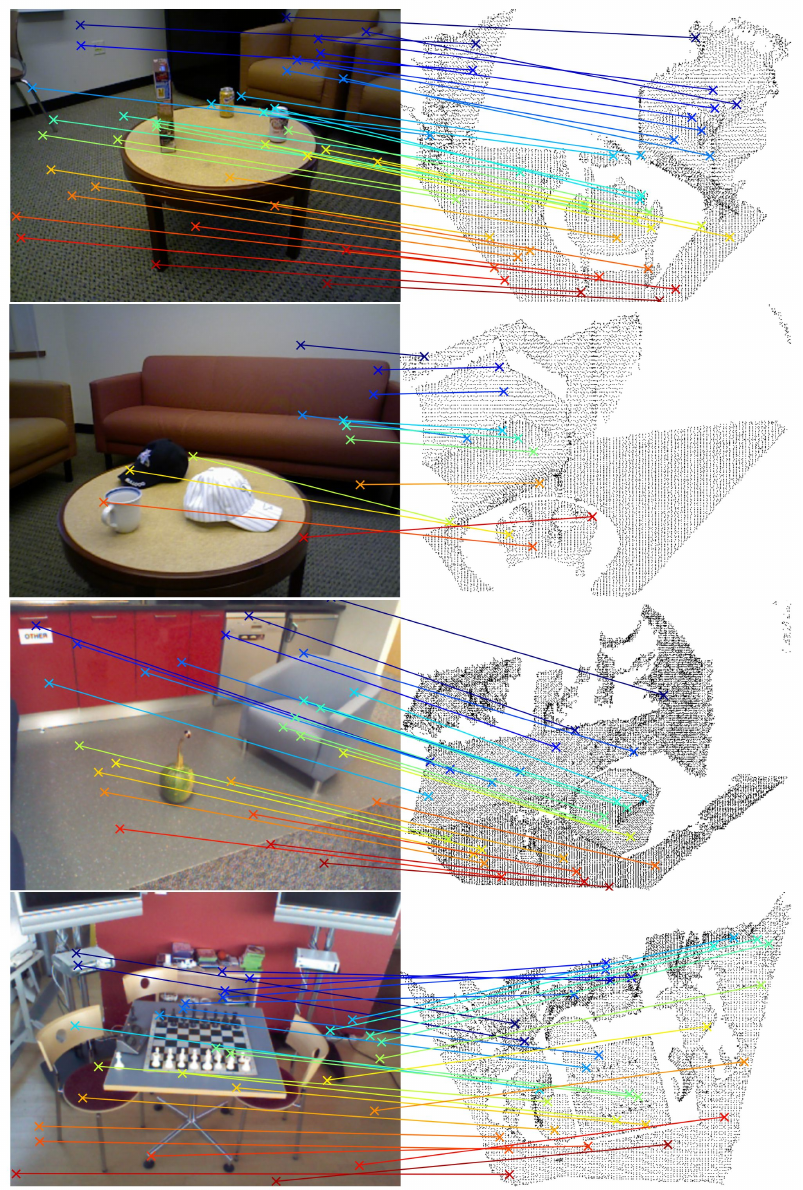}
  \captionof{figure}{\textbf{Visual results of 2D-3D matching on 3DMatch (top) and 3DLoMatch (bottom)}. The top two rows are from the RGB-Scenes V2~\cite{RGB-DScenesV2} and the bottom two rows are from 7Scenes~\cite{7Scenes}.}
\label{fig:2d3d_visual_results}
}]

\twocolumn[{
    \renewcommand\twocolumn[1][]{#1}%
  \centering
   \includegraphics[width=0.95\linewidth, trim=0 0 0 0, clip]{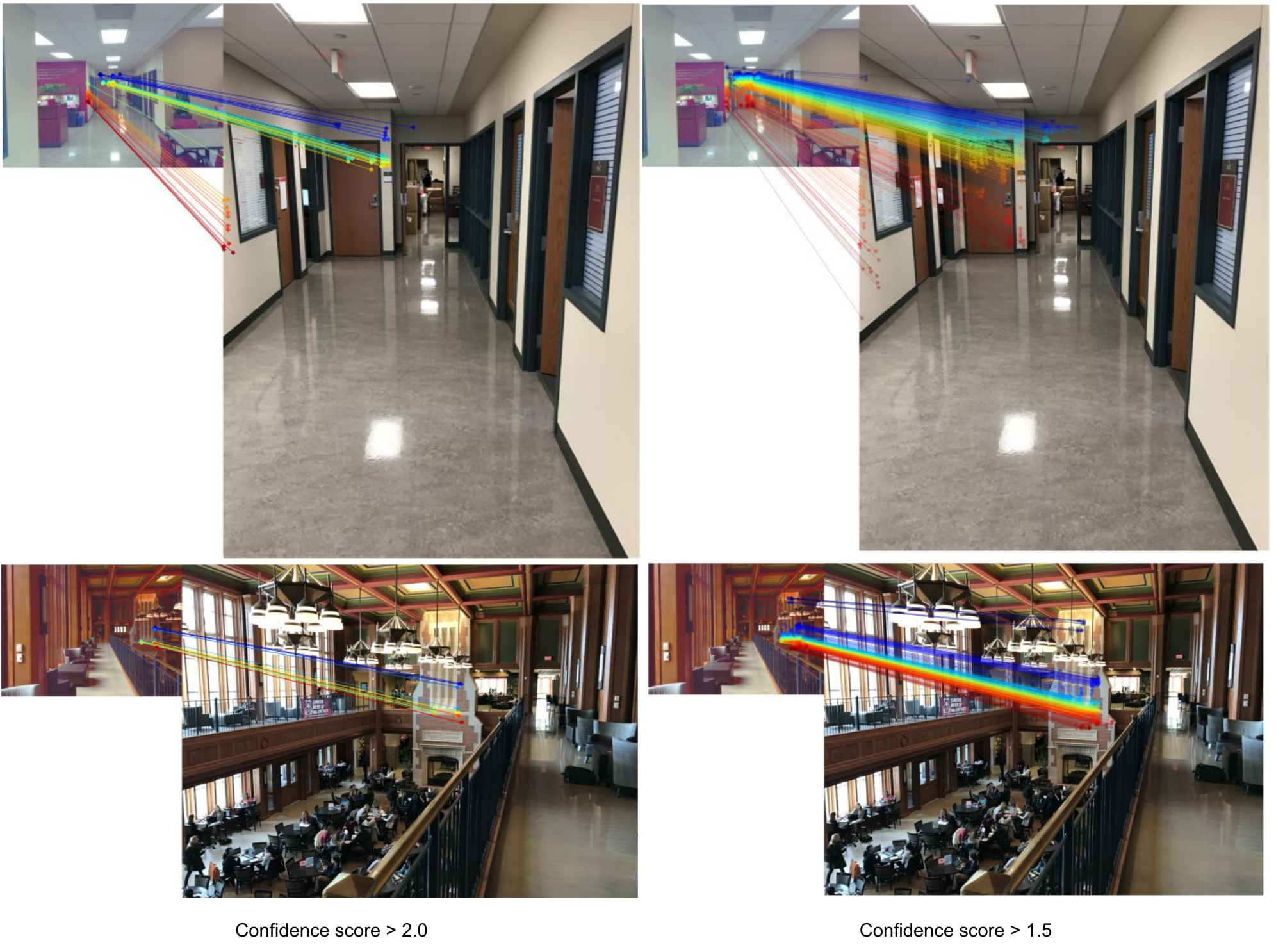}
  \captionof{figure}{\textbf{Visual results on two examples from the InLoc~\cite{InLoc}) Benchmark}. We show the correspondences for different confidence thresholds. Zoom in for details. }
\label{fig:InLoc_qualitative_results}
}]

\begin{figure}[h]
  \centering
   \includegraphics[width=0.95\linewidth, trim=0 20 0 0, clip]{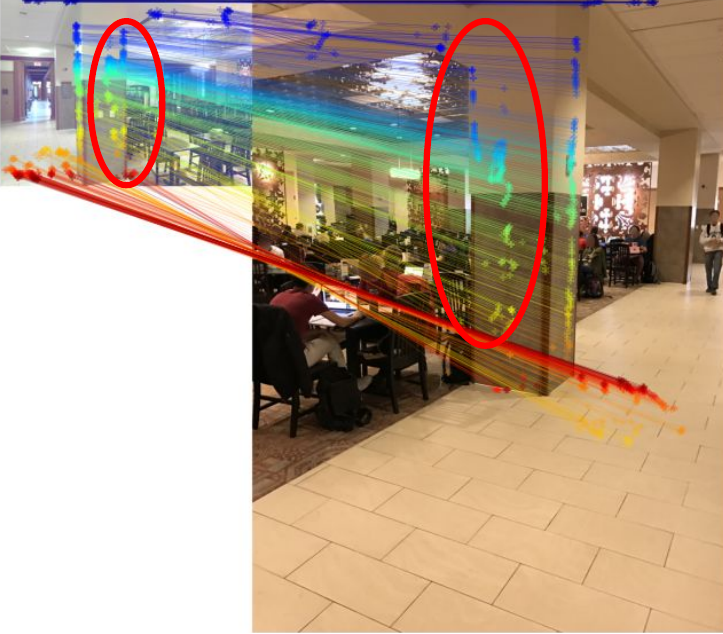}
  \caption{\textbf{Failure case on InLoc~\cite{InLoc} benchmark.} The correspondences inside the red ellipse are invalid since the pillar's face on the first image is not visible on the second image. Hence these correspondences would yield incorrect geometry.}
\label{fig:2d2d_failure_case}
\end{figure}

\begin{table}[htp]
\scriptsize
\setlength{\tabcolsep}{2pt}
    \caption{\textbf{Detailed architecture} of our small-scale and large-scale model.}
    \label{table:suppl_model_details}
    \centering
    \begin{tabular}{l|lll}
    \toprule
    Module              & Type             &  Attribute              & Size     \\
    \midrule
    \multicolumn{4}{l}{\cellcolor{gray!20}\textbf{UniCorrn (small-baseline)}} \\
    Image backbone      & ViT-B~\cite{ViT}    &  Depth                  &  12     \\
                        &             &  Heads                  &  12     \\
                        &                        &  Embedding dims         &  768     \\
    Point cloud backbone      & PTv3~\cite{PTv3} &  Depth                  &  [2, 2, 6, 2]     \\
                              &                  &  Heads                  &  [4, 8, 16, 32]     \\
                              &                  &  Embedding dims         &  [64, 128, 256, 512]     \\
    Feature fusion encoder    & Cross-view ~\cite{CroCov2}   &  Depth                  &  8     \\
                              &                  &  Heads                  &  16     \\
                              &                  &  Embedding dims         &  512     \\
    Matching decoder          &  Dual-stream    &  Depth                   &  8     \\
                              &  (ours)         &  Heads                   &  16     \\
                              &                 &  Embedding dims          &  256     \\
    \midrule
    \multicolumn{4}{l}{\cellcolor{gray!20}\textbf{UniCorrn (small-final)}} \\
    Image backbone      & ViT-B~\cite{ViT}    &  Depth                  &  12     \\
                        &                        &  Heads                  &  12     \\
                        &                        &  Embedding dims         &  768     \\
    Point cloud backbone      & PTv3~\cite{PTv3} &  Depth                  &  [2, 6, 4]     \\
                              &                  &  Heads                  &  [2, 8, 32]     \\
                              &                  &  Embedding dims         &  [32, 128, 512]     \\
    Feature fusion encoder    & Cross-view~\cite{CroCov2}   &  Depth                  &  8     \\
                              &                  &  Heads                  &  16     \\
                              &                  &  Embedding dims         &  512     \\
    Matching decoder          &  Dual-stream    &  Depth                  &  8     \\
                              &  (ours)         &  Heads                  &  1     \\
                              &                 &  Embedding dims         &  256     \\
    \midrule
    \multicolumn{4}{l}{\cellcolor{gray!20}\textbf{UniCorrn (large)}} \\
    Image backbone      & ViT-L~\cite{ViT}   &  Depth                  &  24     \\
                        &                  &  Heads                  &  16     \\
                        &                  &  Embedding dims         &  1024     \\
    Point cloud backbone      & PTv3~\cite{PTv3} &  Depth                  &  [3, 6, 6]     \\
                              &                  &  Heads                  &  [2, 8, 32]     \\
                              &                  &  Embedding dims         &  [32, 128, 512]     \\
    Feature fusion encoder    & Cross-view~\cite{CroCov2}   &  Depth                  &  12     \\
                              &                  &  Heads                  &  16     \\
                              &                  &  Embedding dims         &  768     \\
    Matching decoder          &  Dual-stream     &  Depth                  &  8     \\
                              &  (ours)          &  Heads                  &  1     \\
                              &                  &  Embedding dims         &  256     \\
    \bottomrule
    \end{tabular}
\end{table}   

\begin{table}[htp]
\footnotesize
    \caption{\textbf{Hyper-parameters} for large-scale stage 1 and stage 2 training.}
    \label{table:suppl_training_hparams}
    \centering
    \begin{tabular}{l|ll}
    \toprule
    Hyperparameters         & Stage 1         & Stage 2  \\
    \midrule
    Tasks                   & 2D-2D, 3D-3D            & 2D-2D, 2D-3D, 3D-3D \\
    \midrule
    Optimizer               &  AdamW~\cite{AdamW}     & AdamW~\cite{AdamW} \\
    Base learning rate      &   1e-4   & 2e-5  \\
    Minimum learning rate   &   1e-7      & 1e-7  \\
    Weight decay            &   0.01      & 0.01  \\
    Adam $\beta$            &   (0.9, 0.95) & (0.9, 0.95)  \\
    Batch size (per task)   &  24  &  16 \\
    Epochs                  &  40  & 30 \\
    Warmup epochs           &  4  &  5 \\
    Learning rate scheduler &  Cosine decay        &   Cosine decay        \\
    Gradient norm clipping       &  1.0 &  1.0 \\
    \midrule
    Pre-trained weights     & CroCo V2~\cite{CroCov2} &  Ours (Stage 1)   \\
    \bottomrule
    \end{tabular}
\end{table}   

\begin{table}[ht]
\footnotesize
    \caption{Dataset sample sizes for large-scale training.}
    \label{table:suppl_dataset}
    \centering
    \begin{tabular}{l|ll}
    \toprule
    Dataset                 & Type              & Pairs per epoch     \\
    \midrule
    \multicolumn{3}{l}{\cellcolor{gray!20}\textbf{2D-2D (stage 1)}} \\
    ArkitScenes~\cite{arkitscenes}       & Indoor / Real              & 45,600 \\
    BlendedMVS~\cite{BlendedMVS}         & Mixed / Synthetic          & 68,400 \\
    CO3Dv2~\cite{CO3D}                   & Object-centric / Real      & 22,800 \\
    MegaDepth~\cite{megadepth}           & Outdoor / Real             & 68,400 \\
    Static Things 3D~\cite{things3d}     & Object / Synthetic         & 22,800 \\
    ScanNet++~\cite{scannetpp}           & Indoor / Real              & 60,000 \\
    Waymo~\cite{waymo}                   & Outdoor / Real             & 60,000 \\
    \multicolumn{3}{l}{\cellcolor{gray!20}\textbf{3D-3D (stage 1)}} \\
    3DMatch~\cite{3dmatch}               & Indoor / Real              & 20,586 \\
    ModelNet~\cite{modelnet}             & Object-centric / Synthetic & 5,112 \\
    ArkitScenes~\cite{arkitscenes}       & Indoor / Real              & 80,000 \\
    MegaDepth~\cite{megadepth}           & Outdoor / Real             & 80,000 \\
    ScanNet++~\cite{scannetpp}           & Indoor / Real              & 80,000 \\
    \midrule
    \multicolumn{3}{l}{\cellcolor{gray!20}\textbf{2D-2D (stage 2)}} \\
    MegaDepth~\cite{megadepth}           & Outdoor / Real             & 20,000 \\
    ScanNet++~\cite{scannetpp}           & Indoor / Real              & 20,000 \\
    \multicolumn{3}{l}{\cellcolor{gray!20}\textbf{2D-3D (stage 2)}} \\
    7Scenes~\cite{7Scenes}               & Indoor / Real              & 4,048 \\
    RGB-D Scenes V2~\cite{RGB-DScenesV2} & Indoor / Real              & 1,748 \\
    ScanNet++~\cite{scannetpp}           & Indoor / Real              & 10,000 \\
    \multicolumn{3}{l}{\cellcolor{gray!20}\textbf{3D-3D (stage 2)}} \\
    3DMatch~\cite{3dmatch}               & Indoor / Real              & 20,586 \\
    ModelNet~\cite{modelnet}             & Object-centric / Synthetic & 5,112 \\
    ArkitScenes~\cite{arkitscenes}       & Indoor / Real              & 10,000 \\
    ScanNet++~\cite{scannetpp}           & Indoor / Real              & 20,000 \\    
    \bottomrule
    \end{tabular}
\end{table}   

\begin{table}[bt]
\scriptsize
\setlength{\tabcolsep}{3pt}
    \caption{
    Evaluation results on RGB-D Scenes V2~\cite{RGB-DScenesV2}.
    \textbf{Boldfaced} numbers highlight the best and the second best are \underline{underlined}.
    }
    \label{table:results-rgbdv2}
    \centering
    \begin{tabular}{l|ccccc}
    \toprule
    Model & Scene-11 & Scene-12 & Scene-13 & Scene-14 & Mean \\
    \midrule
    Mean depth (m) & 1.74 & 1.66 & 1.18 & 1.39 & 1.49 \\
    \midrule
    \multicolumn{6}{c}{\emph{Inlier Ratio(IR)} $\uparrow$} \\
    \midrule
    {FCGF-2D3D~\cite{FCGF}} & 6.8 & 8.5 & 11.8 & 5.4 & 8.1 \\
    {P2-Net~\cite{P2-net}} & 9.7 & 12.8 & 17.0 & {9.3} & 12.2 \\
    {Predator-2D3D~\cite{Predator}} & 17.7 & 19.4 & 17.2 & 8.4 & 15.7 \\
    {2D3D-MATR~\cite{2D3D-MATR}} & 32.8 & 34.4 & 39.2 & 23.3 & 32.4 \\
    {B2-3Dnet~\cite{Bridge-2D3D}} & 36.4 & 32.7 & \underline{43.8} & \underline{27.4} & \underline{35.1} \\ 
    {FreeReg~\cite{FreeReg}} & \underline{36.6} & \underline{34.5} & 34.2 & 18.2 & 30.9 \\
    \textbf{Ours (stage 2)} & \textbf{85.7} & \textbf{86.7} & \textbf{92.5} & \textbf{69.1} & \textbf{83.6} \\
    \midrule

    \multicolumn{6}{c}{\emph{Feature Matching Recall (FMR)} $\uparrow$} \\
    \midrule
    {FCGF-2D3D~\cite{FCGF}} & 11.1 & 30.4 & 51.5 & 15.5 & 27.1 \\
    {P2-Net~\cite{P2-net}} & 48.6 & 65.7 & {82.5} & {41.6} & 59.6 \\
    {Predator-2D3D~\cite{Predator}} & {86.1} & {89.2} & 63.9 & 24.3 & {65.9} \\
    {2D3D-MATR~\cite{2D3D-MATR}} & \underline{98.6} & \underline{98.0} & 88.7 & 77.9 & 90.8 \\
    {B2-3Dnet~\cite{Bridge-2D3D}} & \textbf{100.0} & \textbf{99.0} & \underline{92.8} & \underline{85.8} & \underline{94.4} \\ 
    {FreeReg~\cite{FreeReg}} & 91.9 & 93.4 & {93.1} & {49.6} & {82.0} \\ 
    \textbf{Ours (stage 2)} & \underline{98.6} & 97.2 & \textbf{100.0} & \textbf{92.0} & \textbf{97.0} \\
    \midrule

    \multicolumn{6}{c}{\emph{Registration Recall (RR)} $\uparrow$} \\
    \midrule
    {FCGF-2D3D~\cite{FCGF}} & 26.4 & 41.2 & 37.1 & 16.8 & 30.4 \\
    {P2-Net~\cite{P2-net}} & 40.3 & 40.2 & {41.2} & {31.9} & {38.4} \\
    {Predator-2D3D~\cite{Predator}} & 44.4 & 41.2 & 21.6 & 13.7 & 30.2 \\
    {2D3D-MATR~\cite{2D3D-MATR}} & 63.9 & 53.9 & 58.8 & 49.1 & 56.4 \\
    {B2-3Dnet~\cite{Bridge-2D3D}} & 58.3 & 60.8 & 74.2 & 60.2 & 63.4 \\
    {FreeReg~\cite{FreeReg}} & 74.2 & 72.5 & 54.5 & 27.9 & 57.3 \\
    {Diff-Reg~\cite{diffregv2}} & \underline{95.8} &  \textbf{96.1} & \underline{88.7} & \underline{69.0} & \underline{87.4} \\
    \textbf{Ours (stage 2)}     & \textbf{98.6} & \underline{95.3} & \textbf{99.0} & \textbf{76.9} & \textbf{92.5} \\
    \bottomrule
    \end{tabular}
\end{table}

\begin{table}[h]
\scriptsize
\setlength{\tabcolsep}{1.5pt}
    \caption{
    Evaluation results on 7Scenes~\cite{7Scenes}.
    \textbf{Boldfaced} numbers highlight the best and the second best are \underline{underlined}.
    }
    \label{table:results-7scenes}
    \centering
    \begin{tabular}{l|cccccccc}
    \toprule
    Model & Chess & Fire & Heads & Office & Pumpkin & Kitchen & Stairs & Mean \\
    \midrule
    Mean depth (m) & 1.78 & 1.55 & 0.80 & 2.03 & 2.25 & 2.13 & 1.84 & 1.77 \\
    \midrule
    \multicolumn{9}{c}{\emph{Inlier Ratio (IR)} $\uparrow$} \\
    \midrule
    {FCGF-2D3D~\cite{FCGF}} & 34.2 & 32.8 & 14.8 & 26.0 & 23.3 & 22.5 & 6.0 & 22.8 \\
    {P2-Net~\cite{P2-net}} & {55.2} & {46.7} & 13.0 & {36.2} & {32.0} & {32.8} & 5.8 & {31.7} \\
    {Predator-2D3D~\cite{Predator}} & 34.7 & 33.8 & {16.6} & 25.9 & 23.1 & 22.2 & 7.5 & 23.4 \\
    {2D3D-MATR~\cite{2D3D-MATR}} & 72.1 & 66.0 & 31.3 & 60.7 & 50.2 & 52.5 & 18.1 & 50.1 \\
    {B2-3Dnet~\cite{Bridge-2D3D}}  & 73.8 & 66.7 & 33.1 & 61.7 & 50.8 & 52.3 & 18.1 & 50.9 \\
    Diff-Reg~\cite{diffregv2} & \underline{79.2} & \underline{71.0} & \underline{54.1} & \underline{70.4} & \underline{55.8} & \underline{60.2} & \underline{22.9} &  \underline{59.1} \\        
    {Ours (stage 2)} & \textbf{93.7} & \textbf{91.2} & \textbf{92.3} & \textbf{94.8} & \textbf{80.5} & \textbf{87.3} & \textbf{36.7} &  \textbf{82.4} \\
    \midrule
    
    \multicolumn{9}{c}{\emph{Feature Matching Recall (FMR)} $\uparrow$} \\
    \midrule
    {FCGF-2D3D~\cite{FCGF}} &  \underline{99.7} & 98.2 & 69.9 & 97.1 & 83.0 & {87.7} & 16.2 & 78.8 \\
    {P2-Net~\cite{P2-net}} &  \textbf{100.0} & {99.3} & 58.9 & \underline{99.1} & {87.2} & 92.2 & 16.2 & {79.0} \\
    {Predator-2D3D~\cite{Predator}} & 91.3 & 95.1 & {76.7} & 88.6 & 79.2 & 80.6 & {31.1} & 77.5 \\
    {2D3D-MATR~\cite{2D3D-MATR}} & \textbf{100.0} & \underline{99.6} & \underline{98.6} & \textbf{100.0} & \underline{92.4} & {95.9} & {58.1} & {92.1} \\
    {B2-3Dnet~\cite{Bridge-2D3D}}  & \textbf{100.0} & \textbf{100.0} & \underline{98.6} & \textbf{100.0} & \textbf{92.7} & {95.6} & \textbf{64.9} & \textbf{93.1} \\
    Diff-Reg~\cite{diffregv2} & \textbf{100.0} & \textbf{100.0} & \textbf{100.0} & \textbf{100.0} & {91.3} & \underline{98.1} & {58.1} &  {92.5} \\       
    {Ours (stage 2)} & \textbf{100.0} & \textbf{100.0} & \textbf{100.0} & \textbf{100.0} & 90.5 & \textbf{99.9} & \underline{60.7} &  \underline{93.0} \\
    \midrule

    \multicolumn{9}{c}{\emph{Registration Recall (RR)} $\uparrow$} \\
    \midrule
    {FCGF-2D3D~\cite{FCGF}} & {89.5} & 79.7 & 19.2 & 85.9 & 69.4 & 79.0 & 6.8 & 61.4 \\
    {P2-Net~\cite{P2-net}} & {96.9} & {86.5} & {20.5} & {91.7} & {75.3} & {85.2} & 4.1 & {65.7} \\
    {Predator-2D3D~\cite{Predator}} & 69.6 & 60.7 & 17.8 & 62.9 & 56.2 & 62.6 & {9.5} & 48.5 \\
    {2D3D-MATR~\cite{2D3D-MATR}} & {96.9} & {90.7} & {52.1} & {95.5} & {80.9} & {86.1} & {28.4} & {75.8} \\
    {B2-3Dnet~\cite{Bridge-2D3D}}  & \underline{98.3} & {90.5} & {56.2} & {96.4} & \underline{84.0} & {86.1} & \underline{32.4} & {77.7} \\
    Diff-Reg~\cite{diffregv2} & \textbf{100.0} & \underline{94.0} & \underline{90.4} & \underline{99.3} & {81.2} & \underline{94.6} & {27.0} &  \underline{83.8} \\        
    {Ours (stage 2)} & \textbf{100.0} & \textbf{99.3} & \textbf{98.6} & \textbf{100.0} & \textbf{88.8} & \textbf{98.5} & \textbf{51.9} &  \textbf{91.0} \\
    \bottomrule
\end{tabular}
\end{table}







\end{document}